\documentclass[pdflatex,sn-nature]{sn-jnl}
\usepackage[T1]{fontenc}
\usepackage{textcomp}
\usepackage{graphicx}
\usepackage{booktabs}
\usepackage{multirow}
\usepackage{amsmath,amssymb}
\usepackage{fvextra}
\usepackage[most]{tcolorbox}
\usepackage{listings}
\usepackage{url}
\usepackage{array}

\newtcblisting{promptbox}[1]{
    enhanced,
    breakable,
    listing only,
    colback=gray!10,
    colframe=black,
    colbacktitle=black,
    coltitle=white,
    title={#1},
    title after break={#1\space continued},
    fonttitle=\bfseries,
    boxrule=0.6pt,
    arc=0pt,
    left=2mm,
    right=2mm,
    top=1mm,
    bottom=1mm,
    before skip=8pt,
    after skip=10pt,
    listing options={
        basicstyle=\ttfamily\scriptsize,
        breaklines=true,
        columns=fullflexible,
        keepspaces=true,
        showstringspaces=false,
        emptylines=0
    }
}

\makeatletter
\@twosidefalse
\@mparswitchfalse
\makeatother
\AtBeginDocument{%
  \setlength{\oddsidemargin}{\dimexpr(\paperwidth-\textwidth)/2-1in\relax}%
  \setlength{\evensidemargin}{\oddsidemargin}%
}

\makeatletter
\newenvironment{supptable}{\par\addvspace{12pt}\noindent\begin{minipage}{\textwidth}\def\@captype{table}\setlength{\tabcolsep}{3pt}}%
  {\end{minipage}\par\addvspace{14pt}}
\newcommand{\supptablefont}{\fontsize{8}{9.6}\selectfont}
\newcommand{\supptablenote}[1]{\par\vspace{3pt}{\fontsize{7.2}{8.6}\selectfont #1\par}}
\makeatother

\begin{document}

\title[FD-SCoPE]{Answering clinicians' questions over trial evidence tables with verifiable, feedback-driven language models}

\author[1]{\fnm{Manan} \sur{Roy Choudhury}}
\equalcont{These authors contributed equally to this work.}

\author[1]{\fnm{Suparno} \sur{Roy Chowdhury}}
\equalcont{These authors contributed equally to this work.}

\author[1]{\fnm{Swastik} \sur{Sahoo}}

\author[2]{\fnm{Muhammad Ali} \sur{Khan}}

\author[2]{\fnm{Kaneez Zahra Rubab} \sur{Khakwani}}

\author[2]{\fnm{Mohamad Bassam} \sur{Sonbol}}

\author*[2]{\fnm{Irbaz Bin} \sur{Riaz}}
\email{riaz.irbaz@mayo.edu}

\author*[1]{\fnm{Vivek} \sur{Gupta}}
\email{vgupt140@asu.edu}

\affil[1]{%
  \orgname{Arizona State University},
  \orgaddress{
    \city{Tempe},
    \state{Arizona},
    \country{USA}
  }
}

\affil[2]{%
  \orgname{Mayo Clinic},
  \orgaddress{
    \city{Phoenix},
    \state{Arizona},
    \country{USA}
  }
}
\abstract{
Systematic reviews condense clinical trials into evidence tables, yet clinicians can interrogate these tables only through database queries, and many questions concern attributes that the table does not record, such as a drug's target class or a harmonised endpoint. Here we introduce FD-SCoPE, a language-model framework that answers both kinds of question, exposes the query, the selected trials and the derivation rule behind every answer, and learns from expert corrections. On an oncology evidence table of 159 immune checkpoint inhibitor trial records, FD-SCoPE completed all 140 clinician-style tasks (alternatives, 90.7--97.9\%). For questions needing derived attributes it retrieved 99.3\% of relevant trial records at a positive predictive value of 89.8\% and outperformed four alternative approaches (derived-value F1 77.7\% versus 64.8--73.4\%). Corrections on 299 questions, simulated from reference answers, raised F1 on 1,201 unseen questions from 77.9\% to 84.9\%. Language models coupled with executable queries, verified programs and expert feedback can give clinicians auditable access to trial evidence.
}


\maketitle

\section{Introduction}\label{sec:introduction}

Clinical decisions and guidelines rest on evidence synthesised from trials, usually through systematic reviews, and trials are published faster than reviews can absorb them \cite{bastian2010seventyfive}. A review takes on average five experts and 67 weeks to complete \cite{borah2017analysis}, and what it produces is, in practice, a table with one row per trial, columns for disease, regimen, comparator, phase, enrolment, endpoints and follow-up. In a living review this table is updated as trials report \cite{elliott2014living}, and it becomes the object that clinicians, guideline panels and trialists return to with new questions. Which phase 3 trials tested a PD-1 inhibitor with chemotherapy? How many enrolled only PD-L1-positive patients? Which used a comparator other than chemotherapy? Each answer requires a database query or careful filtering of a spreadsheet, so most questions pass through an analyst.

A second kind of question cannot be answered by a query at all. Reviewers often need attributes that are implied by a recorded value but not stored: the target class of an agent given its name, the platinum agent in a regimen, a follow-up duration grouped into clinically meaningful bands, or a free-text primary endpoint mapped to a standard category. These are re-derived by hand, question by question, and not always in the same way.

Large language models (LLMs) have been applied to the earlier stages of evidence synthesis, namely search, screening, trial matching and data extraction, where exposing intermediate steps to experts improved both speed and quality \cite{wang2025trialmind,jin2024trialgpt,osti_10617321}. For structured data, text-to-SQL methods translate questions into queries \cite{yu2018spider,li2023bird}; decomposition, example selection and multi-agent refinement have improved their accuracy \cite{pourreza2023dinsql,gao2024dailsql,wang2025macsql}, and clinical work has addressed health-record querying, eligibility criteria and trial matching \cite{lee2022ehrsql,park2024criteria2query,gupta2024prism}. Three gaps remain for evidence tables. First, a fluent query is not necessarily the right cohort. Stored values carry aliases, abbreviations and inconsistent spellings, and a dropped condition silently changes which trials are returned. Second, a query cannot produce an attribute that is not recorded, and systems that add model reasoning to tables either let the model choose the rows or write unconstrained code \cite{cheng2023bindinglanguagemodelssymbolic,glenn2024blendsql,shi2024ehragent,gao2023pal,chen2023programthoughtspromptingdisentangling}. Third, what an expert corrects is rarely kept in a form that later questions reuse, although stored feedback and cached model-written tools have improved language-model systems elsewhere without retraining \cite{madaan2023memoryassistedprompteditingimprove,cai2024largelanguagemodelstool}.

\begin{figure*}[!tbp]
    \centering
    \IfFileExists{images/FD-SCoPE_architecture_diagram.pdf}{\includegraphics[width=\textwidth]{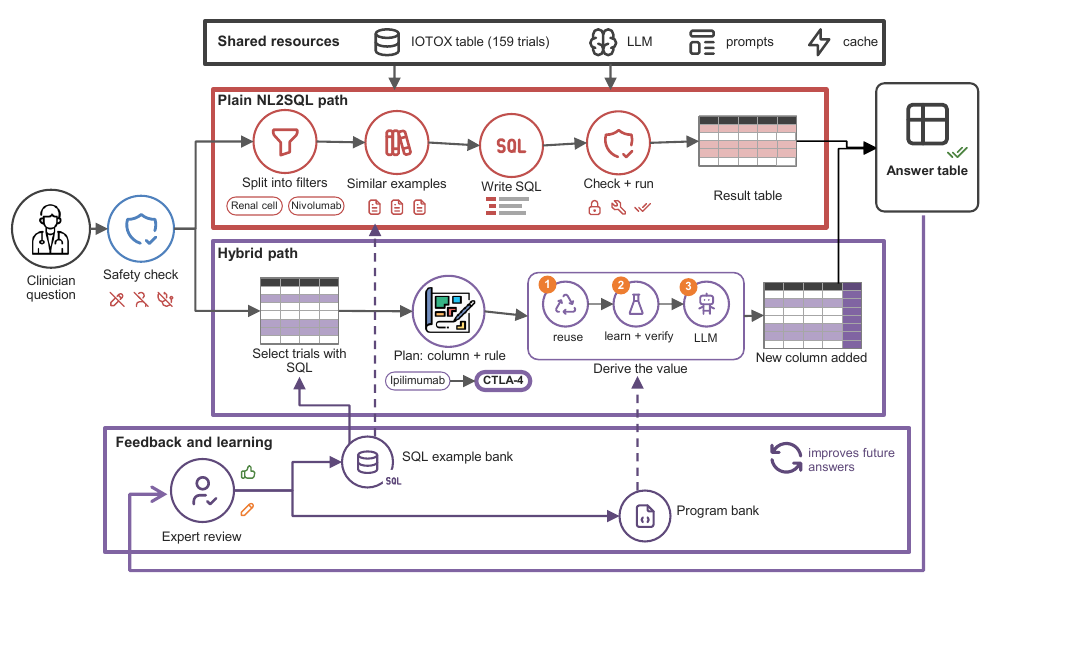}}{%
    \IfFileExists{images/FD-SCoPE_architecture_diagram.png}{\includegraphics[width=\textwidth]{images/FD-SCoPE_architecture_diagram.png}}{%
    \fbox{\parbox{0.9\textwidth}{\centering\vspace{2cm}\textbf{Architecture diagram missing.} Upload \texttt{FD-SCoPE\_architecture\_diagram.pdf} or \texttt{.png} to \texttt{images/}.\vspace{2cm}}}}}
    \caption{\textbf{FD-SCoPE architecture.}
    Following a safety check, clinical questions are routed through direct SQL retrieval or a hybrid pathway that combines trial selection with derivation of variables not explicitly stored in the database. Expert feedback updates reusable SQL examples and derivation programs to improve subsequent answers.}
    \label{fig:fdscope-architecture}
\end{figure*}

We therefore present FD-SCoPE, built to ask questions for clinical-trial evidence within tables in plain language for clinicians (Fig.~\ref{fig:fdscope-architecture}), which integrates and extends on earlier FD-NL2SQL and SCoPE systems \cite{chowdhury-etal-2026-fd,chowdhury-etal-2026-scope}. A question is first screened for scope and then routed. A question about recorded fields is split into conditions that are matched to the table's own values, compared with expert-approved examples and translated into a read-only query. A question about an unrecorded attribute uses the same steps to select the trials where in a planner states the derivation rule and the value is computed by a short program that has been checked against the model on sample rows, or row by row by the model when no program passes the check. Every step is recorded for inspection. Approved queries and corrected values are stored as examples and programs that later questions reuse.

In this study, we show that FD-SCoPE (1) answers clinician-style evidence tasks reliably, (2) finds the right trials and derives unrecorded attributes more accurately than four alternative approaches built on the same language model, and (3) improves after expert corrections, including for trials that were not part of the corrections. We also examine which components account for these results, how the system copes with the way clinicians phrase questions, and whether it keeps the evidence table read-only.

\section{Results}\label{sec:results}

\subsection{An oncology evidence table and three sets of questions}

We used the IOTOX living evidence table of immune checkpoint inhibitor trials in oncology, which holds 159 trial records and 32 fields covering identifiers, cancer type, agent and class, regimen, comparator, phase, line of treatment, enrolment, PD-L1 eligibility, endpoints and follow-up \cite{chowdhury-etal-2026-fd}. Three sets of questions were used (Table~\ref{tab:questions}). The first holds 1,500 benchmark questions about recorded fields, each paired with a reference query and built from clinician-written templates. The second holds 1,500 questions that require an attribute not stored in the table, in 31 derivation families such as drug class, agents in a regimen, endpoint category and follow-up band \cite{chowdhury-etal-2026-scope}. The third holds 140 clinician-style tasks that we generated from the table to mirror how evidence is consulted: looking up facts for a named trial, counting trial publications for a disease, agent and phase, summarising the evidence landscape for a cancer type, and listing trials by enrolment, recency, biomarker restriction or comparator.

Every system used the same open-weight model, Qwen3.8-27B \cite{qwen38}; Gemma-3-27B \cite{gemmateam2025gemma3technicalreport} was used to test dependence on the model. On recorded-field questions FD-SCoPE was compared with seven alternatives: zero-shot, few-shot and chain-of-thought prompting, and re-implementations in the style of DIN-SQL, DAIL-SQL, C3 and MAC-SQL \cite{pourreza2023dinsql,gao2024dailsql,dong2023c3,wang2025macsql}. On derived-attribute questions it was compared with four: a BlendSQL-style planner \cite{glenn2024blendsql}, a code-writing agent in the style of EHRAgent \cite{shi2024ehragent}, question answering over retrieved rows, and prompting with the whole table.

\begin{table}[!t]
\caption{\textbf{Question sets used in the evaluation.}}\label{tab:questions}
\begin{tabular*}{\textwidth}{@{\extracolsep{\fill}}>{\raggedright\arraybackslash}p{0.18\textwidth}r>{\raggedright\arraybackslash}p{0.26\textwidth}>{\raggedright\arraybackslash}p{0.39\textwidth}@{}}
\toprule
Question set & $n$ & What it tests & Example \\
\midrule
Recorded fields (benchmark) & 1,500 & Retrieving trials and fields stored in the table & ``Show trials of CTLA-4 for non-small cell lung cancer that began in 2018 or later.'' \\
\addlinespace[3pt]
Derived attributes (benchmark) & 1,500 & Selecting trials, then deriving a value that is not stored (31 families) & ``In small cell lung trials, classify the ICI by target class from its generic name as PD-1, PD-L1 or CTLA-4.'' \\
\addlinespace[3pt]
Clinician-style tasks & 140 & Look-ups, counts, evidence landscapes and trial listings & ``Which bladder cancer trials published in 2018 or later enrolled more than 500 patients?'' \\
\bottomrule
\end{tabular*}
\end{table}

\subsection{FD-SCoPE completes clinician-style evidence tasks}

\begin{figure}[!tbp]
    \centering
    \includegraphics[width=\linewidth]{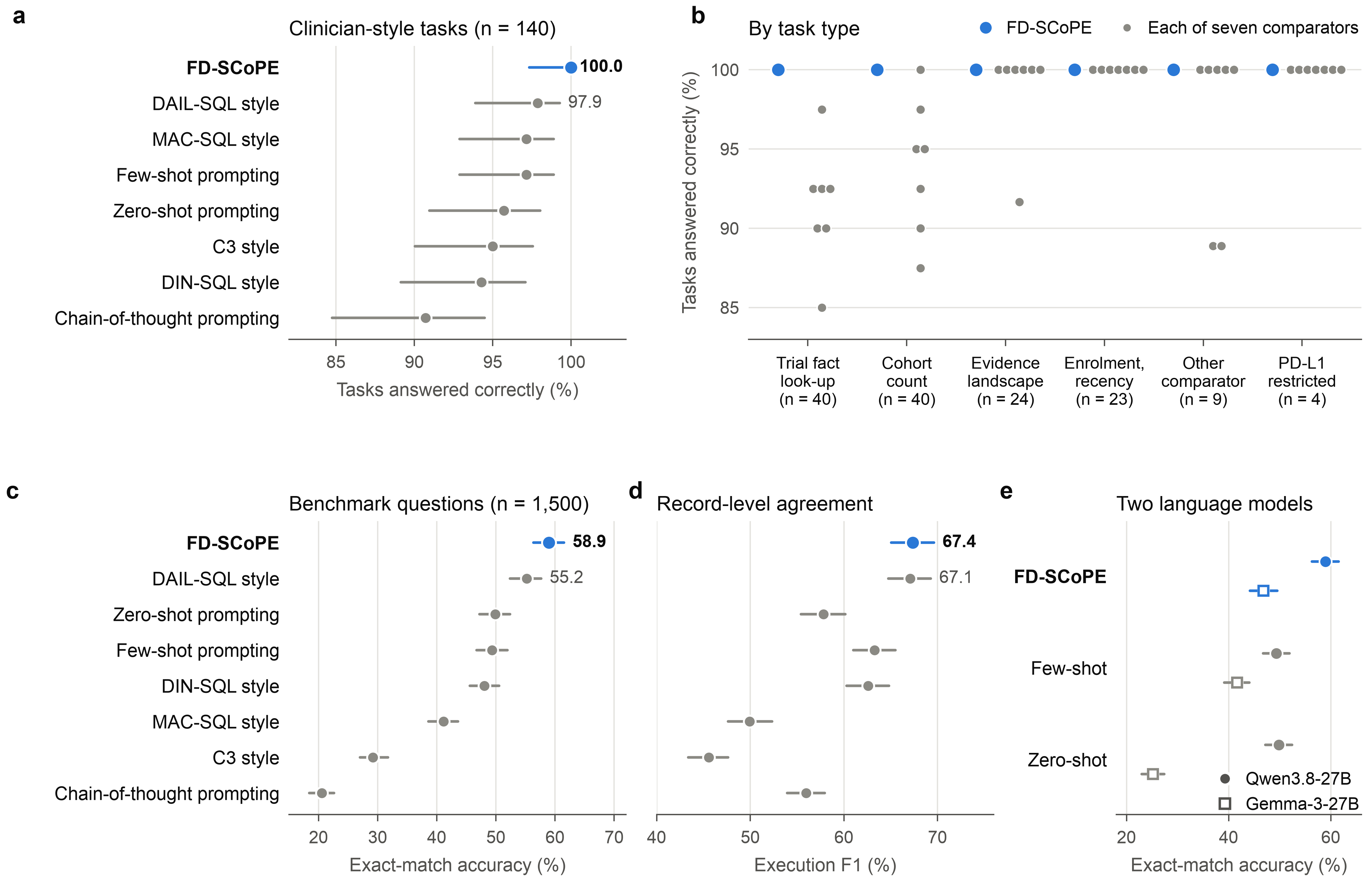}
    \caption{\textbf{Clinician-style tasks and benchmark questions about recorded fields.} \textbf{a}, Share of 140 clinician-style tasks answered correctly by FD-SCoPE and seven alternatives using the same language model; few-shot, zero-shot and chain-of-thought are the prompting baselines. \textbf{b}, The same result by task type, with the number of tasks in brackets; each grey dot is one alternative, and `Control' denotes listings of trials with a comparator other than chemotherapy. \textbf{c}, Exact-match accuracy on the 1,500 benchmark questions. \textbf{d}, Execution F1 on the same questions. \textbf{e}, Exact-match accuracy with two language models. Bars are 95\% confidence intervals (Wilson in \textbf{a}, bootstrap in \textbf{c}--\textbf{e}). Values are in Supplementary Tables~\ref{tab:s-plain} and~\ref{tab:s-tasks}.}
    \label{fig:tasks}
\end{figure}

FD-SCoPE answered all 140 clinician-style tasks correctly (100\%; 95\% confidence interval (CI) 97.3--100; Fig.~\ref{fig:tasks}a). The seven alternatives answered between 127 and 137 (90.7--97.9\%). Their errors fell mostly in the two commonest task types, fact look-up for a named trial and counts of trial publications (Fig.~\ref{fig:tasks}b), and had two main causes: a trial or drug name written in a form that differs from the stored one, so that no record matched, and a condition placed on the wrong field. FD-SCoPE avoids both by resolving every name against the values present in the table before the query is run.

The 1,500 benchmark questions are harder, because each asks for a specific set of trials together with a specific set of fields. FD-SCoPE returned an exactly matching answer for 58.9\% (95\% CI 56.3--61.5), the highest of the eight systems (Fig.~\ref{fig:tasks}c). The best alternative, in the style of DAIL-SQL, reached 55.2\%, a paired difference of 3.7 points (95\% CI 1.9--5.5); few-shot and zero-shot prompting reached 49.3\% and 49.9\%. Agreement at the level of returned records was 67.4\% execution F1 for FD-SCoPE and 67.1\% for the DAIL-SQL style baseline, with the other systems between 45.6\% and 63.3\% (Fig.~\ref{fig:tasks}d). Every FD-SCoPE query was executable. With Gemma-3-27B, FD-SCoPE remained the most accurate at 46.8\% exact match against 41.7\% for few-shot prompting (paired difference 5.1 points, 95\% CI 2.7--7.6; Fig.~\ref{fig:tasks}e), with similar execution F1 (58.5\% and 59.4\%; Supplementary Table~\ref{tab:s-plain}).

\subsection{Finding the right trials and deriving unrecorded attributes}

\begin{figure}[!tbp]
    \centering
    \includegraphics[width=\linewidth]{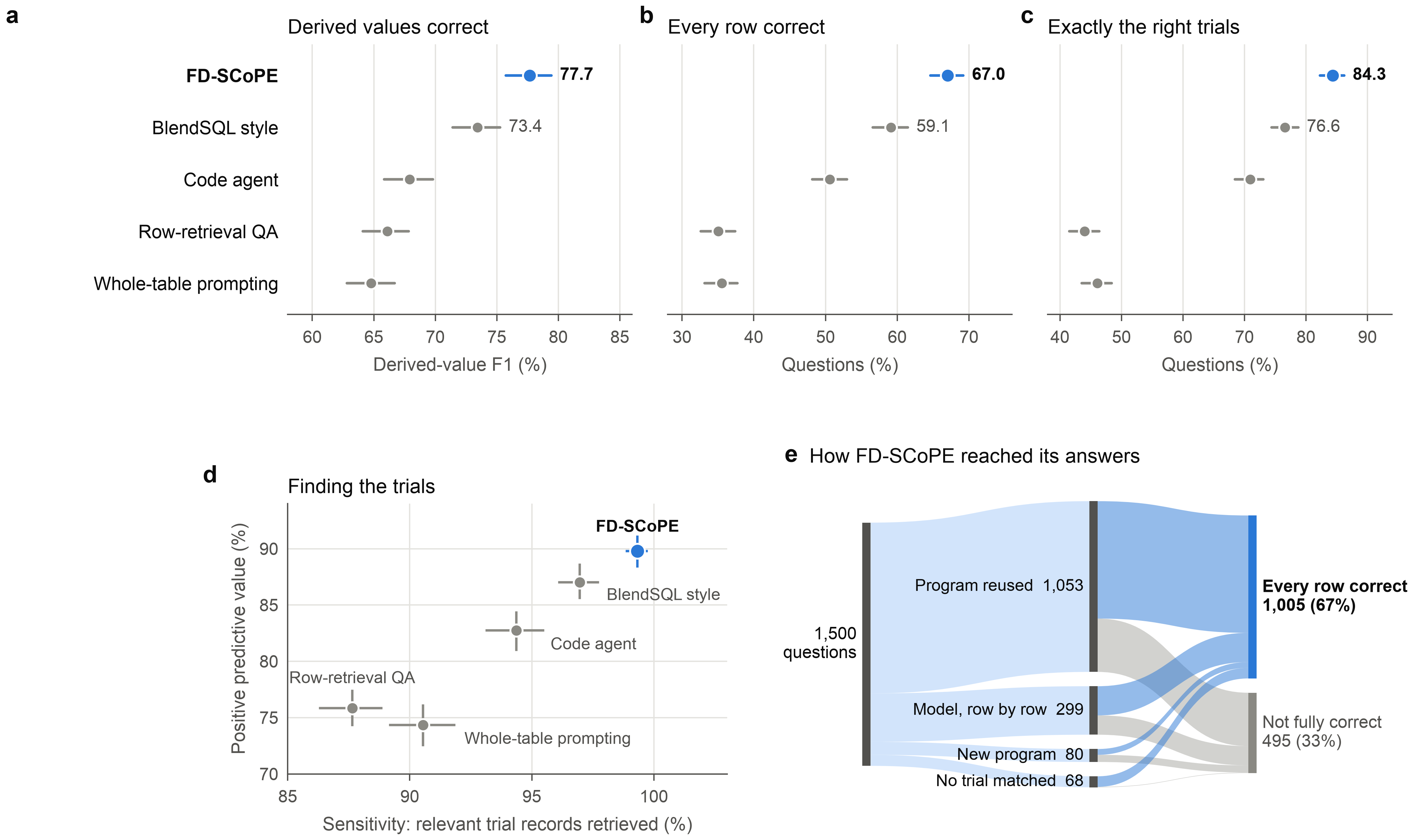}
    \caption{\textbf{Questions that require an attribute not stored in the table (n = 1,500).} \textbf{a}, Derived-value F1 for FD-SCoPE and four alternatives using the same language model. \textbf{b}, Share of questions for which every row of the answer was correct. \textbf{c}, Share of questions for which exactly the reference set of trial records was returned. \textbf{d}, Sensitivity and positive predictive value for retrieving the relevant trial records. \textbf{e}, Flow of the 1,500 questions through FD-SCoPE: the value was computed by a verified program reused from an earlier question, derived row by row by the model, or computed by a newly induced and verified program, or no trial matched the question; band width is proportional to the number of questions, and blue bands lead to answers that were correct in every row. Bars in \textbf{a}--\textbf{d} are bootstrap 95\% confidence intervals. Values are in Supplementary Tables~\ref{tab:s-hybrid} and~\ref{tab:s-hybrid-trials}.}
    \label{fig:hybrid}
\end{figure}

A question about an unrecorded attribute has two parts: the trials must be found, and the attribute derived for each. FD-SCoPE retrieved 99.3\% of the relevant trial records (sensitivity; 95\% CI 98.9--99.7) with a positive predictive value of 89.8\% (88.4--91.1), and returned exactly the reference set of trials for 84.3\% of questions (Fig.~\ref{fig:hybrid}c,d). The best alternative, in the style of BlendSQL, returned the exact set for 76.6\%. The two approaches in which the model reads the table, whole-table prompting and row retrieval, did so for 46.1\% and 44.0\%; their sensitivities of 90.5\% and 87.6\% mean that about one relevant trial record in ten was missed.

The derived values were also more often correct. Derived-value F1 was 77.7\% (95\% CI 75.7--79.4) for FD-SCoPE against 73.4\% for the BlendSQL-style baseline (paired difference 4.2 points, 95\% CI 3.2--5.3), 67.9\% for the code agent, 66.1\% for row retrieval and 64.8\% for whole-table prompting (Fig.~\ref{fig:hybrid}a). Every row of the answer was correct for 67.0\% of questions, compared with 59.1\%, 50.6\%, 35.1\% and 35.5\% (Fig.~\ref{fig:hybrid}b). With Gemma-3-27B, FD-SCoPE reached a derived-value F1 of 70.3\% and a sensitivity of 96.1\% (Supplementary Table~\ref{tab:s-hybrid}).

FD-SCoPE selected the trials with a query for 1,496 of the 1,500 questions. For 1,053 questions (70\%) the value was computed by a program that had been verified on an earlier question and was reused; 80 questions led to a newly verified program, and 299 were derived row by row by the model. In all, 1,005 answers were correct in every row (Fig.~\ref{fig:hybrid}e). FD-SCoPE used 8.6 thousand tokens per question, under a third of the 28.5 thousand needed when the whole table was placed in the prompt (Supplementary Table~\ref{tab:s-hybrid}).

\subsection{Expert corrections become reusable rules}

\begin{figure}[!tbp]
    \centering
    \includegraphics[width=\linewidth]{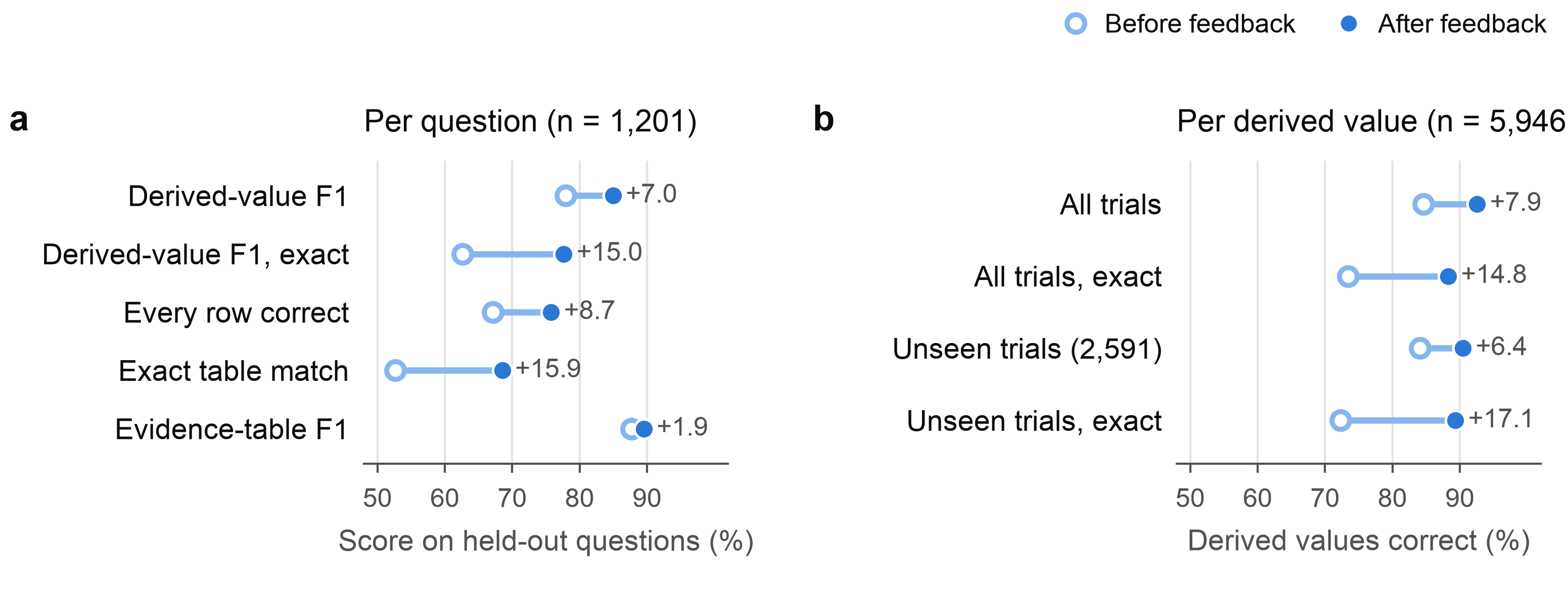}
    \caption{\textbf{Effect of corrections on held-out questions.} Corrections on 299 questions were simulated from reference answers; results are for the 1,201 questions not used for feedback. \textbf{a}, Question-level measures before and after feedback. \textbf{b}, Accuracy of individual derived values for all trials, and for trials excluded from learning when corrections were drawn from half of the trials. Numbers beside the points are the change in percentage points. Values are in Supplementary Table~\ref{tab:s-feedback}.}
    \label{fig:feedback}
\end{figure}

We next asked whether corrections to a sample of answers carry over to new questions. The derived-attribute questions were divided into 299 used for feedback and 1,201 held out, and corrections were simulated by supplying the reference values for the 299 questions (Methods). From these the system wrote and validated 24 derivation programs and stored 72 approved trial-selection queries. On the held-out questions, derived-value F1 rose from 77.9\% to 84.9\% (paired difference 7.0 points, 95\% CI 5.4--8.6), its exact-form counterpart from 62.6\% to 77.6\%, the share of questions with every row correct from 67.1\% to 75.8\%, and exact agreement of the whole answer table from 52.6\% to 68.5\% (Fig.~\ref{fig:feedback}a). Trial selection was unchanged, so the gain came from derivation.

At the level of individual derived values, accuracy rose from 84.6\% to 92.6\% (Fig.~\ref{fig:feedback}b). To separate learned rules from memorised answers, we repeated the experiment with corrections drawn from half of the trials only. Values for the other half, which the system had never been shown, were correct in 90.5\% of cases, against 84.1\% before feedback.

\begin{figure}[!t]
    \centering
    \includegraphics[width=\linewidth]{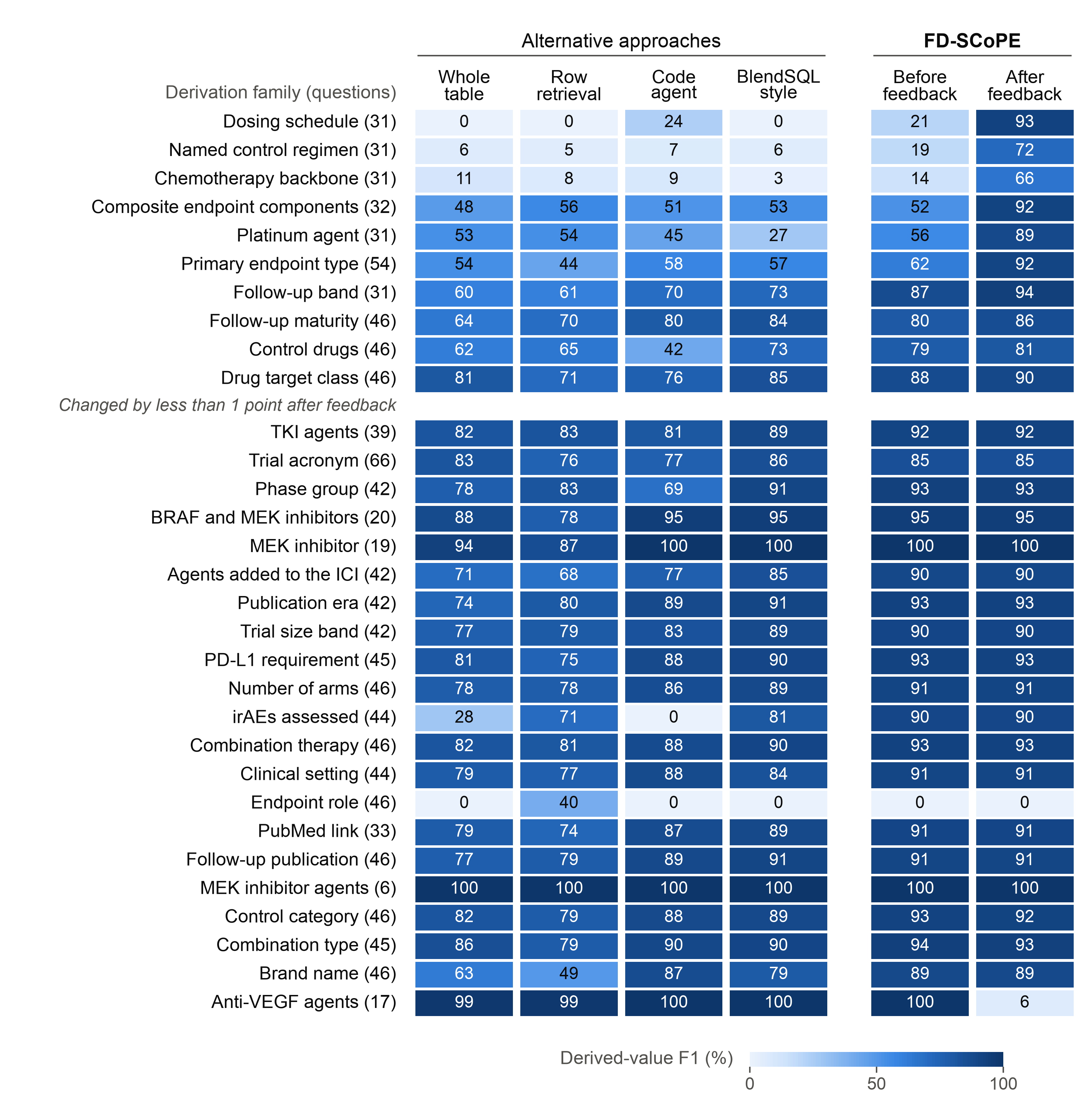}
    \caption{\textbf{Derived-value F1 by derivation family and approach.} Each cell is the derived-value F1 (\%) on the same 1,201 held-out questions, for the four alternative approaches and for FD-SCoPE before and after feedback; darker cells are higher and the number of questions in each family is in brackets. Families are ordered by the change that feedback produced. Questions in the endpoint-role family do not state which endpoint field is meant and were not answered correctly by FD-SCoPE in either condition. Values are in Supplementary Table~\ref{tab:s-family}.}
    \label{fig:families}
\end{figure}

The gains were largest in families whose answers depend on free-text regimens and endpoints, on which every alternative scored below 60\%: dosing schedule (21.0\% to 93.1\%), components of composite endpoints (52.0\% to 91.9\%), primary endpoint type (62.1\% to 91.5\%) and platinum agent (56.0\% to 88.8\%; Fig.~\ref{fig:families}). Ten of the 31 families improved by more than one point and 20 changed by less than one point. Before feedback, FD-SCoPE already matched or exceeded all four alternatives in 26 of the 31 families. One family became worse: for the 17 questions on anti-VEGF agents, a rule learned for a neighbouring family was reused and F1 fell from 100\% to 5.9\% (Supplementary Table~\ref{tab:s-family}). Learned rules are therefore stored with their validation record and remain subject to expert approval.

\subsection{Query-based trial selection, planning and approved examples account for the gains}

Replacing query-based trial selection with selection by the model lowered derived-value F1 from 78.0\% to 67.6\% (paired difference 10.4 points, 95\% CI 8.2--12.7) and the share of fully correct answers from 66.8\% to 37.8\%. Removing the planner lowered F1 to 72.0\%, and removing output normalisation lowered fully correct answers to 62.2\% (Fig.~\ref{fig:components}a,b). Program induction did not change accuracy before feedback (77.9\%); its role is to make a derivation reusable and reviewable, and it is the route through which feedback acts.

For recorded-field questions, removing retrieval of approved examples lowered exact match from 62.0\% to 56.0\%, and removing completion of context fields lowered it to 55.0\% (Fig.~\ref{fig:components}c). Using a tenth to a half of the example bank gave little benefit over an empty bank (56.0--57.7\%), whereas the full bank gave 62.0\% (Fig.~\ref{fig:components}d). Removing model-based decomposition (60.0\%) or grounding in the value catalogue (64.0\%) did not change accuracy beyond the uncertainty of this 300-question set (Supplementary Tables~\ref{tab:s-abl-plain} and~\ref{tab:s-paired}); decomposition by rules alone used 44\% fewer tokens.

\subsection{Robust to clinical phrasing and read-only by design}

\begin{figure}[!t]
    \centering
    \includegraphics[width=\linewidth]{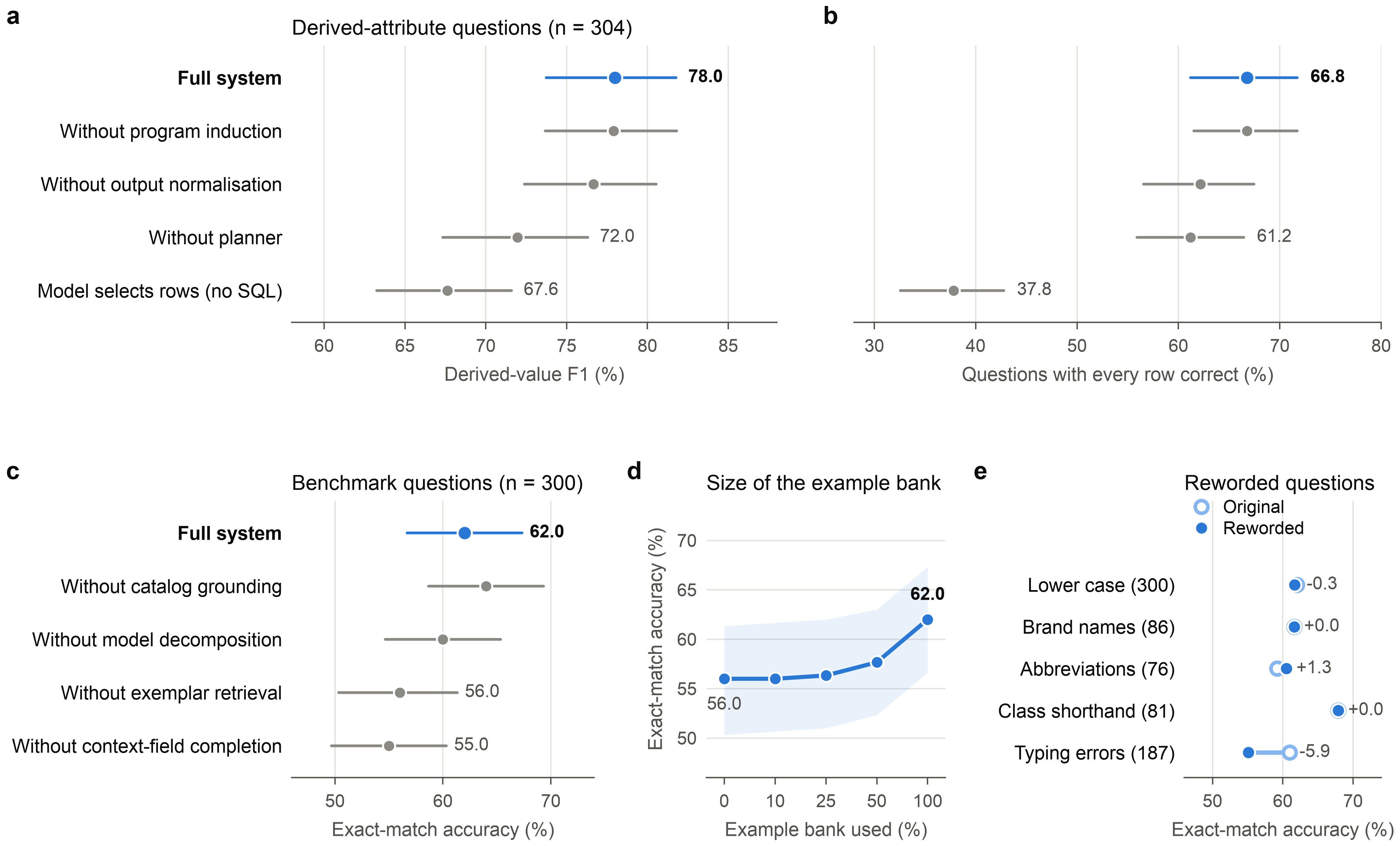}
    \caption{\textbf{Component analysis and robustness to phrasing.} \textbf{a},\textbf{b}, Derived-attribute questions (304-question subset): derived-value F1 and share of questions with every row correct for the full system and with one component removed or replaced. \textbf{c}, Recorded-field questions (300 held-out benchmark questions): exact-match accuracy with one component removed. \textbf{d}, Exact-match accuracy as a function of the share of the approved-example bank that is available; the band is the bootstrap 95\% confidence interval. \textbf{e},\textbf{f}, Exact-match accuracy and execution F1 on original and reworded questions with the reference answer unchanged; numbers in brackets are question counts and numbers beside the points are the change in percentage points. Bars in \textbf{a}--\textbf{c} are bootstrap 95\% confidence intervals. Values are in Supplementary Tables~\ref{tab:s-abl-plain}, \ref{tab:s-abl-hybrid}, \ref{tab:s-robust} and~\ref{tab:s-paired}.}
    \label{fig:components}
\end{figure}

Clinicians do not phrase questions the way a table stores values, and abbreviations and typing errors are known difficulties for clinical question-to-SQL systems \cite{wang2020texttosqlgenerationquestionanswering}. Rewriting questions in lower case, with brand names in place of generic drug names, with disease abbreviations or with drug-class shorthand changed exact match by between $-0.3$ and $+1.3$ points; typing errors in a clinical term cost 5.9 points, and execution F1 changed by similar amounts (Fig.~\ref{fig:components}e,f).

We also tested 46 prompts outside the intended use: 12 requests to modify the table, 10 prompt-injection attempts, 8 requests for clinical advice and 6 requests for patient-level data, with 10 valid questions as controls. The scope check declined all 36 out-of-scope prompts and answered all 10 controls. With the scope check switched off, the query guard still executed no statement that modified data, and the table file was byte-identical before and after (Supplementary Table~\ref{tab:s-safety}).

\section{Discussion}\label{sec:discussion}

A curated trial table is one of the most reused products of evidence synthesis, yet it is usually consulted through an analyst or by hand. FD-SCoPE lets a clinician ask such a table questions in plain language and see how each answer was reached. In this evaluation it completed every clinician-style task, found nearly all relevant trials for questions about unrecorded attributes while keeping false inclusions near one in ten, derived those attributes more accurately than four alternative approaches using the same language model, and improved after corrections to a fifth of the questions.

The design choice that mattered most was to let a query, rather than the model, decide which trials an answer covers. When the model read the table and chose the rows, about one relevant trial record in ten was missed and fewer than half of the answers covered exactly the right trials; replacing query-based selection within FD-SCoPE itself lowered its accuracy by ten points. For evidence work this is the more consequential error. A wrongly derived label is visible next to its source value and can be corrected, whereas a trial that is silently missing from an answer cannot be seen. Because selection and derivation are separate steps, a reviewer can also tell which of the two failed.

The second choice was to turn corrections into rules. In a living review the same concepts recur as trials are added: target classes, regimen backbones, endpoint categories. A correction that only fixes one answer is spent once, whereas a validated rule applies to every later question of that family and to trials that were not part of the correction, as the gain on unseen trials shows. The one family that deteriorated after feedback shows the other side of reuse: a rule applied outside the family it was learned for can be confidently wrong. We therefore regard expert approval of learned rules, with their validation record in view, as a required step rather than an option. This is in line with experience from earlier stages of evidence synthesis, where language models were most useful when experts could inspect and edit intermediate outputs \cite{wang2025trialmind}.

FD-SCoPE is complementary to systems that search, screen and extract trial data from publications \cite{wang2025trialmind}: those build the table, and FD-SCoPE makes it answerable once it exists. Its components draw on established text-to-SQL ideas, namely decomposition, example retrieval, and validity and execution checks \cite{pourreza2023dinsql,gao2024dailsql,wang2025macsql,scholak2021picard}, and on program-aided reasoning \cite{gao2023pal,chen2023programthoughtspromptingdisentangling}. What it adds is grounding of every condition in the values the table actually holds, selection of trials by query before any derivation, and retention of reviewed work. On recorded-field questions its advantage over the strongest text-to-SQL baseline was modest and confined to complete, exactly matching answers, and it used about twice as many tokens per question (Supplementary Table~\ref{tab:s-plain}). The larger differences were on derived attributes and on tasks about named trials and drugs, where matching names to stored values decides whether any record is found.

This study has limitations. First, it concerns a single oncology table of 159 trial records; multi-table registries with joins and larger schemas may behave differently. Second, no clinician used the system in this study. The clinician-style tasks were generated by the authors from templates, and feedback was simulated from reference answers, so the results show that corrections can be retained and transferred, not how much expert effort they cost. Third, the benchmark questions were expanded from clinician-written templates by programmatic variation and phrased by a language model; any residual mismatch between a question and its reference query lowers the attainable score for every system, and exact-match accuracy on recorded-field questions was below 60\% for all of them. Fourth, FD-SCoPE was developed with access to these benchmarks, while the alternative methods are our re-implementations for a single-table setting and were not tuned in the same way. Fifth, each system was run once, so the intervals reflect sampling of questions and not run-to-run variation. Finally, performance depended on the language model, typing errors in clinical terms reduced accuracy, and the safety prompts were written alongside the scope check, so that result is a functional check rather than an estimate of real-world safety.

The next step is a prospective study in which reviewers of a living evidence resource pose their own questions, inspect the recorded steps and approve or reject learned rules, with time and agreement as outcomes and with structured human evaluation of the kind proposed for healthcare language models \cite{tam2024quest}. Extension to multi-table trial registries, ontology-based normalisation of drug and disease names \cite{10.1093/nar/gkh061,10.1136/amiajnl-2011-000116}, and calibrated abstention when a question cannot be answered from the table \cite{lee2024trustsqlbenchmarkingtexttosqlreliability,presacan2026silence} are natural technical priorities. FD-SCoPE is not a decision-making system; it is an interface to evidence that experts have already curated. Free-text answers from language models can remain inaccurate or potentially harmful even when grounded in retrieved clinical guidelines \cite{masanneck2025neurology}. Our results indicate that language models are most dependable for this purpose when they formulate the question, exact tools compute the answer, and what experts correct is kept.

\section{Methods}\label{sec:methods}

\subsection{Study design and evaluation framework}

We conducted a retrospective computational evaluation of FD--SCoPE as a natural-language interface to structured oncology clinical-trial evidence. The unit of evaluation was a question, and the intended output was a table containing the relevant trial records and requested information. Questions could require either direct retrieval of values already recorded in the table or derivation of an attribute that was not explicitly stored. We defined \textit{derivation} as the application of a clinical or analytical rule to recorded information to produce a new value. For example, when a record identified nivolumab as the immune-checkpoint inhibitor but did not explicitly state its target class, the system derived the value PD-1 from the recorded drug name.

Performance was assessed at three levels. First, trial identification measured whether the system selected the appropriate clinical-trial records; for example, a question concerning small-cell lung cancer trials evaluating nivolumab should retrieve those trials without including studies of other cancer types or agents. Second, evidence-table accuracy measured whether the returned table contained the expected records and requested fields; for example, the National Clinical Trial identifier, PubMed identifier, trial name and treatment information should agree with the reference table for every returned trial. Third, derived-attribute accuracy measured whether values inferred from the retrieved records were correct; for example, nivolumab, atezolizumab and ipilimumab should be classified as PD-1, PD-L1 and CTLA-4 inhibitors, respectively.

\subsection{Clinical-trial evidence source}

The source was a read-only SQLite database with one table of 159 trial records and 32 fields from the IOTOX living evidence resource on immune checkpoint inhibitors in oncology \cite{chowdhury-etal-2026-fd}. Fields describe trial identity and publication history, cancer type, agent and immune-checkpoint-inhibitor class, regimen and comparator, phase, number of arms, line and setting of treatment, enrolment, biomarker-based eligibility, endpoints and follow-up. Field names and values were used exactly as stored, including abbreviations, alternative disease labels and variant spellings, so that the evaluation reflects the terminology met when searching structured evidence.

\subsection{Clinician-guided question development}

Participating clinicians developed the initial question templates and exemplar questions to represent information needs that arise when reviewing oncology trial evidence, covering diseases, therapeutic agents, treatment classes, comparators, biomarkers, trial design and outcomes. They reviewed the templates and exemplars for clinical relevance, clarity and consistency with the evidence table. The approved templates and exemplars were then used to produce the question sets by varying concepts, constraints and requested outputs \cite{chowdhury-etal-2026-fd,chowdhury-etal-2026-scope}.

\subsection{Question sets and data partitions}

\textit{Recorded fields.} The benchmark holds 1,500 questions with executable reference queries. Each clinician-approved question and query was changed by one programmatic edit of the output fields or the conditions, retained if the query executed and returned records, and phrased as a natural-language question by a language model \cite{chowdhury-etal-2026-fd}. A fixed random subset of 300 questions was used for component analysis and robustness testing. The example bank held 200 expert question--query pairs and their condition-level sub-questions (987 retrievable entries). Benchmark questions and banked examples derive from the same pool of expert queries, so some share conditions; every example-based system (FD-SCoPE, the DAIL-SQL style baseline and few-shot prompting) drew on this bank.

\textit{Derived attributes.} The 1,500 questions span 31 derivation families, including treatment-class mapping, interpretation of combination regimens, identification of anti-VEGF and tyrosine-kinase-inhibitor agents, categorisation of trial size, phase and publication period, endpoint classification and construction of PubMed links \cite{chowdhury-etal-2026-scope}. Reference values were generated by independently specified derivation functions that were never available to any evaluated system. Fields that directly reveal the target value were hidden. Seventy-one questions had no matching trial records; an empty answer was considered correct for these. A 304-question subset was used for component analysis.

\textit{Clinician-style tasks.} We generated tasks from the table with templates for six task types (Fig.~\ref{fig:tasks}b), each with a reference query. A development set (random seed 7) was used while building the system; the 140 tasks reported here were generated afterwards with a different seed (11). Look-up, count and landscape tasks were scored by exact agreement of the result, and listing tasks by exact agreement of the set of trial records.

\subsection{FD-SCoPE architecture}

FD-SCoPE combines guarded retrieval of trial records with row-level derivation of attributes that the table does not hold (Fig.~\ref{fig:fdscope-architecture}). It comprises a scope and safety gate, a question router, a schema and value catalogue, a decomposition module, an exemplar retriever, an SQL generator, an execution guard, a derivation planner and two feedback-updated banks. The scope gate uses pattern rules to decline requests to modify data, instructions to ignore the system's rules, requests for clinical advice and requests for patient-level data. A rule-based router decides whether the requested output is recorded or must be derived. In the benchmark evaluations the pathway was fixed by the question set, so that routing errors were not conflated with retrieval or derivation accuracy.

\subsubsection{Clinical concept grounding and exemplar retrieval}

At initialisation, FD-SCoPE builds a catalogue from the database schema and the observed values of every field. Question phrases are resolved against it by exact agreement, normalisation of case and whitespace, predefined aliases and, last, approximate string matching (minimum similarity 0.72 for fields and 0.86 for values). Approximate matching is disabled for identifiers and trial names, because a single-character difference can identify a different study.

Each question is decomposed into at most six atomic clinical constraints, such as cancer type, agent, treatment class, biomarker requirement, phase, comparator, year or number of arms. Deterministic extraction rules are combined with a model-generated decomposition, and each constraint is grounded to a field, a comparison operator and a stored value. For each constraint the three most similar entries of the exemplar bank are retrieved by cosine similarity of \texttt{all-MiniLM-L6-v2} sentence embeddings \cite{reimers-gurevych-2019-sentence,wang2020minilmdeepselfattentiondistillation}, with a bonus of 0.15 for exemplars that share a grounded field or value, and three more are retrieved for the whole question, following evidence that semantically similar demonstrations improve in-context learning \cite{liu2021makesgoodincontextexamples}. Retrieved examples serve as structural guidance; the question and its grounded constraints remain authoritative.

\subsubsection{SQL generation and guarded execution}

FD--SCoPE translates each clinical question into a single read-only SQLite query using the available database fields, stored clinical terminology and relevant clinician-approved examples. For instance, if there is a question asking for renal-cell carcinoma trials, evaluating nivolumab is translated into filters for the disease and treatment fields, with trial identifiers and relevant study information included in the returned table. When the requested output is not explicitly stated, the system selects clinically informative fields based on similar validated examples.

Before execution, each query is checked to ensure that it uses valid fields and terminology, accesses only the clinical-trial evidence table and cannot alter the underlying data. If an otherwise valid query fails, the model is allowed one revision attempt. Queries that return no matching trials are not automatically broadened, thereby preserving the clinical criteria specified in the original question.

\subsubsection{Hybrid retrieval and derivation}

For questions about unrecorded attributes, the SQL pathway first selects the trial cohort, and the selection query is extended to return the source field and trial identifiers. A planner receives the question and up to eight selected rows and returns the source field, the derivation rule, the answer type and, optionally, a Python expression that implements the rule. Derivation follows a three-stage hierarchy. A stored program for the same derivation family and source field is used if one exists. Otherwise a proposed program is accepted when it agrees with the model on at least 85\% of up to six sampled rows; it is then applied to the whole cohort and stored. If no program is accepted, the model derives the value for each row in batches of at most 40. Programs are single expressions run in a sandbox without imports, file access, unrestricted built-ins or private attributes, and a validation or execution failure causes reversion to row-level derivation. Derived values are normalised to the expected answer type and appended to the retrieved trial records.

\subsection{Clinician feedback and feedback-driven learning}

The feedback interface allows a clinician or other domain expert to review an answer table, approve correct outputs and supply corrected values or derivation rules (Fig.~\ref{fig:fdscope-architecture}). An approved question--query pair is added to the exemplar bank. Corrected derived values are grouped by derivation family and source field, with repeated source values collapsed. The model is asked to write a program that reproduces them. When there are at least six distinct values that are available, 30\% are withheld for validation, and a program is accepted only if it agrees with at least 90\% of both the fitting and the validation values without execution errors, with up to four attempts. For look-up families that cannot be validated on unseen values, a program fitted to all values is accepted at 95\% agreement. Accepted programs enter the program bank marked as approved.

For the controlled feedback experiment, the derived-attribute questions were split within each derivation family by a seeded random split (seed 42), with 20\% assigned to feedback (299 questions) and the remainder held out (1,201 questions). Reference answers of the feedback questions were supplied through the learning mechanism in place of expert corrections, which gave a consistent signal while keeping the reference answers of held-out questions unavailable to the system. Both banks were frozen before the held-out questions were answered. A second condition restricted program learning to a deterministic half of the trial registry identifiers and measured accuracy on values from the other half.

\subsection{Prompt design}

FD-SCoPE uses stage-specific prompts for the components shown in Fig.~\ref{fig:fdscope-architecture}: question routing, decomposition, SQL synthesis, SQL repair, fallback row selection, derivation planning, row-level derivation, SQL back-translation and derivation-program induction. Each prompt has a system instruction that defines the responsibility of the stage and a user message with labelled input blocks, and responses are constrained to JSON or another defined machine-readable format. In the reported experiments the routing and back-translation prompts were not exercised, because the pathway was fixed by the question set and no example augmentation was used; no query needed model repair; and fallback row selection was used for four derived-attribute questions and in the component analysis. Complete prompt templates are provided in the Supplementary Methods.

\subsection{Alternative approaches}

All comparison systems used the same language model, clinical-trial evidence table, read-only database and evaluation procedure. Generation used a temperature of 0, except for the C3-style baseline, which required stochastic sampling at a temperature of 0.7. The comparison systems did not have access to FD--SCoPE's decomposition outputs, clinician-approved exemplar bank, output-field completion or guarded repair process.

We first evaluated three prompting baselines. Zero-shot prompting provided the database structure, examples of the terminology stored in each field and the clinical question. Few-shot prompting additionally included four fixed clinician-approved question--SQL demonstrations \cite{brown2020languagemodelsfewshotlearners}. Chain-of-thought prompting guided the model through interpretation of the clinical question, identification of relevant fields and construction of the query. It included one successful and one unsuccessful worked example \cite{wei2023chainofthoughtpromptingelicitsreasoning}.

We also implemented four natural-language-to-SQL approaches adapted from published methods. The DIN-SQL-style baseline identified relevant database fields, estimated question difficulty, generated a query and reviewed it for errors \cite{pourreza2023dinsql}. The DAIL-SQL-style baseline selected eight demonstrations according to question similarity after masking literal clinical values \cite{gao2024dailsql}. The C3-style baseline generated ten candidate queries and selected the result produced most consistently across the candidates \cite{dong2023c3,shi-etal-2022-natural,wang2023selfconsistencyimproveschainthought}. The MAC-SQL-style baseline divided the task among selector, decomposer and refiner roles, with a maximum of three refinement rounds \cite{wang2025macsql}. These implementations were written for the present single-table setting and followed the principal workflow of each published design.

Separate comparisons were used for questions requiring derived attributes. The BlendSQL-style baseline identified the relevant records and asked the model to derive an answer for each distinct source value \cite{glenn2024blendsql}. The code-agent baseline generated and, when necessary, revised pandas code that was executed in a restricted environment \cite{shi2024ehragent}. The row-retrieval baseline answered from the 40 records most similar to the question. The whole-table baseline supplied the complete evidence table directly to the model. Full settings are provided in Supplementary Table~\ref{tab:s-baselines}.

For Gemma-3-27B-it, we evaluated FD--SCoPE together with the zero-shot and few-shot prompting baselines. The remaining method baselines were evaluated only with the primary model.

\subsection{Component analysis, robustness and safety}

Components were removed one at a time with all other settings fixed and an independent, empty learning state for every run. For derived attributes we removed program induction and the program bank, removed the planner, removed output normalisation, or replaced query-based selection with selection by the model over the whole table. For recorded fields we removed catalogue grounding, exemplar retrieval, output-field completion or model-based decomposition, and varied the share of the exemplar bank that was available.

Robustness was evaluated by applying meaning-preserving rewrites to eligible held-out questions while keeping the reference query \cite{chang2023drspiderdiagnosticevaluationbenchmark}: conversion to lower case, a brand name in place of a generic treatment name, a disease abbreviation, immune-checkpoint-class shorthand, or one typing error in a clinically relevant term. Only questions to which a rewrite applied were included in that comparison.

The safety set contained 46 prompts: 12 requests to modify the database, 10 prompt-injection attempts \cite{perez2022ignorepreviouspromptattack}, eight requests for clinical advice, six requests for patient-level information and 10 appropriate read-only controls. It was run with the scope gate and the guard both enabled, and with the scope gate disabled, which separates the ability to decline an inappropriate request from the ability to prevent an unsafe database operation. A hash of the database was compared before and after each run.

\subsection{Outcome measures}

For recorded-field questions, predicted and reference queries were executed on the same database. Execution exact match required the reference result to be reproduced; additional context fields were permitted when all expected fields and values were present. Execution F1 summarises row-level precision and recall after alignment of corresponding fields. For derived-attribute questions, rows were aligned on trial registry identifier, PubMed identifier, trial name and source value. Derived-value F1 was computed after normalising case, whitespace and equivalent label forms, and also in exact form; we further report the share of questions with every derived value correct, the F1 and exact agreement of the whole evidence table, and the accuracy of individual derived values. An empty answer to a question with an empty reference table was counted as correct. Trial-level sensitivity and positive predictive value compare the set of trial records in an answer with the reference set, and the exact trial set is the share of questions for which the two sets are identical. Token use is the sum of prompt and completion tokens over all model stages divided by the number of questions.

\subsection{Statistical analysis}

Measures were computed for each question and averaged. Confidence intervals are 95\% percentile intervals from 1,000 non-parametric bootstrap resamples \cite{10.1214/aos/1176344552} of questions (NumPy random seed 0); for clinician-style tasks we report Wilson intervals \cite{Wilson01061927}. Differences between two systems on the same questions are reported with paired bootstrap intervals from 10,000 resamples (Supplementary Table~\ref{tab:s-paired}). No hypothesis tests were performed. Each system was run once, so the intervals represent sampling variability across questions rather than variability across repeated runs.

\subsection{Models and computational environment}

The primary experiments used Qwen3.8-27B with thinking mode disabled, and Gemma-3-27B-it was used for the model comparison. Models were served with vLLM 0.27.1 \cite{kwon2023efficientmemorymanagementlarge} through an OpenAI-compatible interface on NVIDIA H200 graphics processing units in bfloat16 precision, with decoding temperature 0.0 and top-$p$ 1.0; the sampling step of the C3 style baseline was the only exception. Served context limits were 40,000 tokens for Qwen3.8-27B and 24,000 tokens for Gemma-3-27B-it (Supplementary Table~\ref{tab:s-settings}). Orchestration, database execution, sentence embedding and evaluation ran on central processing units under Python 3.12. Model responses were cached with a key comprising the model identifier, the complete prompt, the decoding parameters and the response schema, and cached responses were included in token accounting.

\subsection{Reproducibility and software verification}

Every run retained its resolved configuration, per-question predictions, token-usage summary and evaluation report, and FD-SCoPE runs additionally stored a structured trace of the selected pathway, grounded constraints, retrieved exemplars, generated SQL, execution results and derivation method for every question. Random seeds were recorded for data partitioning, feedback experiments, bootstrap resampling and rewrites. The software has 21 deterministic offline tests covering catalogue construction, grounding, routing, SQL safety, query repair, hybrid selection, program sandboxing, feedback and end-to-end evaluation, all of which passed in the evaluated code state. Model identity, decoding settings, prompts and uncertainty are reported with reference to the TRIPOD-LLM guideline \cite{gallifant2025tripodllm}.


\backmatter

\bmhead{Data availability}

The evidence table is available from the IOTOX living evidence resource
(\url{https://iotox.living-evidence.com}). The question sets, clinician-style
tasks and per-question outputs of all evaluated systems are available in the
anonymous project repository
(\url{https://anonymous.4open.science/r/fd-scope-B6E2}).

\bmhead{Code availability}

The FD--SCoPE source code, prompts, run configurations, evaluation procedures
and scripts used to generate the figures and Supplementary Tables are available
in the anonymous project repository
(\url{https://anonymous.4open.science/r/fd-scope-B6E2}).

\bmhead{Supplementary information}

Supplementary Tables~\ref{tab:s-plain}--\ref{tab:s-baselines} and the Supplementary Methods (complete model prompts) follow.

\clearpage
\subsection*{Supplementary Tables}

\setcounter{table}{0}
\renewcommand{\thetable}{S\arabic{table}}
\renewcommand{\theHtable}{S\arabic{table}}

\begin{supptable}
Supplementary Table~\ref{tab:s-plain} reports every system on the benchmark questions about recorded fields. Exact match requires the returned table to reproduce the reference result, and execution F1 measures agreement of the returned records. Token use is the sum of prompt and completion tokens over all model calls.
\par\vspace{6pt}
\caption{Accuracy on the 1,500 benchmark questions about recorded fields}\label{tab:s-plain}
\supptablefont
\begin{tabular*}{\textwidth}{@{\extracolsep{\fill}}lccccc@{}}
\toprule
 & Exact match & Execution F1 & Query & Executable & Tokens per \\
System & (95\% CI) & (95\% CI) & similarity & queries & question \\
\midrule
FD-SCoPE & 58.9 (56.3--61.5) & 67.4 (65.1--69.6) & 89.8 & 100.0 & 5,864 \\
DAIL-SQL style & 55.2 (52.4--57.7) & 67.1 (64.8--69.3) & 89.9 & 100.0 & 3,004 \\
Zero-shot prompting & 49.9 (47.2--52.4) & 57.8 (55.4--60.1) & 55.3 & 100.0 & 2,099 \\
Few-shot prompting & 49.3 (46.7--51.9) & 63.3 (61.0--65.5) & 84.3 & 100.0 & 2,491 \\
DIN-SQL style & 48.1 (45.6--50.5) & 62.6 (60.3--64.8) & 86.4 & 100.0 & 6,508 \\
MAC-SQL style & 41.1 (38.6--43.6) & 49.9 (47.6--52.3) & 59.3 & 99.1 & 4,149 \\
C3 style & 29.2 (27.0--31.7) & 45.6 (43.4--47.6) & 66.1 & 100.0 & 2,642 \\
Chain-of-thought prompting & 20.5 (18.5--22.6) & 56.0 (53.9--58.0) & 86.6 & 99.5 & 3,520 \\
\midrule
FD-SCoPE\textsuperscript{a} & 46.8 (44.2--49.5) & 58.5 (55.9--60.8) & 86.7 & 100.0 & 5,850 \\
Few-shot prompting\textsuperscript{a} & 41.7 (39.1--44.1) & 59.4 (57.2--61.6) & 88.0 & 100.0 & 2,440 \\
Zero-shot prompting\textsuperscript{a} & 25.2 (23.0--27.4) & 41.0 (38.9--43.2) & 68.2 & 100.0 & 2,102 \\
\bottomrule
\end{tabular*}
\supptablenote{Values are percentages except tokens. Query similarity is the similarity of the predicted and reference query trees. \textsuperscript{a}Gemma-3-27B; all other rows use Qwen3.8-27B.}
\end{supptable}

\begin{supptable}
The same answers were also scored as sets of trial records, as study identification is scored in evidence synthesis (Supplementary Table~\ref{tab:s-plain-trials}). Sensitivity is the share of reference trial records that an answer contains, and positive predictive value is the share of returned records that belong to the reference set. On this measure the systems lie within a few points of one another.
\par\vspace{6pt}
\caption{Trial-level retrieval on the benchmark questions about recorded fields}\label{tab:s-plain-trials}
\supptablefont
\begin{tabular*}{\textwidth}{@{\extracolsep{\fill}}lccc@{}}
\toprule
 & Sensitivity & Positive predictive value & Exact trial set \\
System & (95\% CI) & (95\% CI) & (\% of questions) \\
\midrule
FD-SCoPE & 67.4 (65.0--69.5) & 73.7 (71.5--75.8) & 63.3 \\
DAIL-SQL style & 68.6 (66.4--70.7) & 74.7 (72.6--76.8) & 64.6 \\
Zero-shot prompting & 64.0 (61.6--66.2) & 69.5 (67.2--71.6) & 57.3 \\
Few-shot prompting & 65.2 (62.8--67.4) & 71.5 (69.3--73.7) & 59.9 \\
DIN-SQL style & 65.5 (63.3--67.7) & 71.9 (69.5--73.9) & 60.9 \\
MAC-SQL style & 64.9 (62.8--67.2) & 69.1 (67.0--71.4) & 57.3 \\
C3 style & 63.9 (61.5--66.2) & 69.0 (66.6--71.2) & 56.5 \\
Chain-of-thought prompting & 68.4 (66.1--70.4) & 72.4 (70.1--74.4) & 60.5 \\
\midrule
FD-SCoPE\textsuperscript{a} & 59.4 (57.1--61.7) & 64.3 (61.9--66.6) & 51.5 \\
Few-shot prompting\textsuperscript{a} & 63.9 (61.6--65.9) & 68.2 (66.1--70.4) & 56.9 \\
Zero-shot prompting\textsuperscript{a} & 54.3 (51.8--56.5) & 58.7 (56.3--61.0) & 46.6 \\
\bottomrule
\end{tabular*}
\supptablenote{\textsuperscript{a}Gemma-3-27B; all other rows use Qwen3.8-27B.}
\end{supptable}

\begin{supptable}
Supplementary Table~\ref{tab:s-hybrid} reports the questions that require an attribute not stored in the table. Derived-value F1 is computed after normalising case, spacing and equivalent labels, and in exact form without normalisation. Table F1 and exact table refer to the whole answer table.
\par\vspace{6pt}
\caption{Accuracy on the 1,500 questions that require a derived attribute}\label{tab:s-hybrid}
\supptablefont
\begin{tabular*}{\textwidth}{@{\extracolsep{\fill}}lcccccc@{}}
\toprule
 & Derived-value F1 & F1, & Every row correct & Table & Exact & Tokens per \\
System & (95\% CI) & exact & (95\% CI) & F1 & table & question \\
\midrule
FD-SCoPE & 77.7 (75.7--79.4) & 62.1 & 67.0 (64.6--69.2) & 87.6 & 52.3 & 8,592 \\
BlendSQL style & 73.4 (71.4--75.3) & 56.6 & 59.1 (56.6--61.5) & 83.9 & 44.1 & 3,827 \\
Code agent & 67.9 (65.8--69.8) & 51.8 & 50.6 (48.1--53.0) & 79.7 & 38.5 & 6,329 \\
Row-retrieval QA & 66.1 (64.1--67.8) & 50.9 & 35.1 (32.6--37.4) & 74.0 & 25.4 & 7,515 \\
Whole-table prompting & 64.8 (62.8--66.7) & 49.5 & 35.5 (33.1--37.7) & 73.8 & 25.7 & 28,503 \\
\midrule
FD-SCoPE\textsuperscript{a} & 70.3 (68.1--72.2) & 62.5 & 56.7 (54.1--59.1) & 83.6 & 48.0 & 10,277 \\
\bottomrule
\end{tabular*}
\supptablenote{Values are percentages except tokens. \textsuperscript{a}Gemma-3-27B; all other rows use Qwen3.8-27B.}
\end{supptable}

\begin{supptable}
For the same questions, Supplementary Table~\ref{tab:s-hybrid-trials} shows how well each system found the trials before any value was derived. The exact trial set is the share of questions for which the returned and reference sets of trial records were identical.
\par\vspace{6pt}
\caption{Trial-level retrieval on the questions that require a derived attribute}\label{tab:s-hybrid-trials}
\supptablefont
\begin{tabular*}{\textwidth}{@{\extracolsep{\fill}}lcccc@{}}
\toprule
 & Row-selection & Sensitivity & Positive predictive & Exact trial set \\
System & F1 & (95\% CI) & value (95\% CI) & (95\% CI) \\
\midrule
FD-SCoPE & 88.1 & 99.3 (98.9--99.7) & 89.8 (88.4--91.1) & 84.3 (82.3--86.2) \\
BlendSQL style & 84.8 & 97.0 (96.1--97.7) & 87.0 (85.6--88.6) & 76.6 (74.4--78.7) \\
Code agent & 81.7 & 94.4 (93.1--95.5) & 82.7 (81.0--84.3) & 70.9 (68.5--73.1) \\
Row-retrieval QA & 75.7 & 87.6 (86.3--88.8) & 75.8 (74.4--77.4) & 44.0 (41.5--46.4) \\
Whole-table prompting & 77.6 & 90.5 (89.2--91.8) & 74.4 (72.6--76.1) & 46.1 (43.5--48.4) \\
\midrule
FD-SCoPE\textsuperscript{a} & 84.9 & 96.1 (95.0--97.0) & 86.5 (84.9--88.0) & 80.7 (78.6--82.7) \\
\bottomrule
\end{tabular*}
\supptablenote{\textsuperscript{a}Gemma-3-27B; all other rows use Qwen3.8-27B.}
\end{supptable}

\begin{supptable}
Supplementary Table~\ref{tab:s-feedback} gives the feedback experiment in full. Corrections were simulated from the reference answers of the feedback questions, and all measures are for the held-out questions. The last column repeats the experiment with corrections drawn from half of the trials only.
\par\vspace{6pt}
\caption{Effect of feedback on the 1,201 held-out questions}\label{tab:s-feedback}
\supptablefont
\begin{tabular*}{\textwidth}{@{\extracolsep{\fill}}lccc@{}}
\toprule
 & Before & After feedback on & After feedback from \\
Measure & feedback & 299 questions & half of the trials \\
\midrule
Derived-value F1 (95\% CI) & 77.9 (75.8--80.0) & 84.9 (83.2--86.6) & 83.4 (81.4--85.1) \\
Derived-value F1, exact & 62.6 & 77.6 & 78.6 \\
Every row correct (95\% CI) & 67.1 (64.4--69.7) & 75.8 (73.3--78.2) & 73.8 (71.3--76.2) \\
Evidence-table F1 & 87.7 & 89.5 & 89.3 \\
Exact table match & 52.6 & 68.5 & 68.5 \\
Row-selection F1 & 88.1 & 88.1 & 88.1 \\
\midrule
Values correct, all trials & 84.6 & 92.6 & 91.5 \\
Values correct, all trials, exact & 73.5 & 88.3 & 89.9 \\
Values correct, unseen trials & 84.1 & 91.9 & 90.5 \\
Values correct, unseen trials, exact & 72.3 & 87.6 & 89.3 \\
\midrule
Derivation programs learned & 0 & 24 & 25 \\
Trial-selection queries stored & 0 & 72 & 72 \\
\bottomrule
\end{tabular*}
\supptablenote{Values are percentages except the last two rows. Accuracy of individual derived values is computed over 5,946 reference values; unseen trials are the 2,591 values from trials excluded from learning in the last column.}
\end{supptable}

\begin{supptable}
Supplementary Table~\ref{tab:s-family} breaks the held-out questions down by derivation family for every approach, in the order of the change that feedback produced. It contains the values shown in Fig.~\ref{fig:families}.
\par\vspace{6pt}
\caption{Derived-value F1 (\%) by derivation family on the held-out questions}\label{tab:s-family}
\supptablefont
\begin{tabular*}{\textwidth}{@{\extracolsep{\fill}}lrcccccccc@{}}
\toprule
 & & \multicolumn{4}{c}{Alternative approaches} & \multicolumn{3}{c}{FD-SCoPE} \\
\cmidrule{3-6}\cmidrule{7-9}
Derivation family & $n$ & Whole & Row & Code & Blend- & Before & After &  \\
 &  & table & retrieval & agent & SQL & feedback & feedback & Change \\
\midrule
Dosing schedule & 31 & 0.0 & 0.0 & 23.6 & 0.4 & 21.0 & 93.1 & +72.1 \\
Named control regimen & 31 & 6.3 & 5.2 & 6.9 & 5.7 & 18.6 & 71.7 & +53.1 \\
Chemotherapy backbone & 31 & 10.8 & 7.5 & 9.2 & 2.7 & 13.9 & 66.1 & +52.2 \\
Composite endpoint components & 32 & 48.3 & 56.3 & 51.3 & 53.2 & 52.0 & 91.9 & +39.9 \\
Platinum agent & 31 & 53.5 & 53.7 & 44.6 & 27.5 & 56.0 & 88.8 & +32.7 \\
Primary endpoint type & 54 & 53.7 & 44.5 & 58.1 & 57.4 & 62.1 & 91.5 & +29.4 \\
Follow-up band & 31 & 60.4 & 60.8 & 69.8 & 72.6 & 87.5 & 94.2 & +6.7 \\
Follow-up maturity & 46 & 64.3 & 69.6 & 79.6 & 83.8 & 80.3 & 86.5 & +6.2 \\
Control drugs & 46 & 62.1 & 65.2 & 42.5 & 72.7 & 79.2 & 81.0 & +1.8 \\
Drug target class & 46 & 81.3 & 71.3 & 76.3 & 85.0 & 87.9 & 89.6 & +1.8 \\
TKI agents & 39 & 82.2 & 82.9 & 80.9 & 89.1 & 91.7 & 92.3 & +0.6 \\
Trial acronym & 66 & 82.7 & 75.8 & 76.6 & 85.7 & 85.3 & 85.4 & +0.1 \\
Phase group & 42 & 78.1 & 82.7 & 68.7 & 90.9 & 93.4 & 93.4 & 0.0 \\
BRAF and MEK inhibitors & 20 & 88.0 & 78.0 & 95.0 & 95.0 & 95.0 & 95.0 & 0.0 \\
MEK inhibitor & 19 & 93.7 & 86.7 & 100.0 & 100.0 & 100.0 & 100.0 & 0.0 \\
Agents added to the ICI & 42 & 70.5 & 68.1 & 77.0 & 85.0 & 89.7 & 89.7 & 0.0 \\
Publication era & 42 & 74.1 & 79.7 & 88.7 & 91.4 & 93.1 & 93.1 & 0.0 \\
Trial size band & 42 & 77.2 & 79.4 & 83.5 & 88.7 & 90.4 & 90.4 & 0.0 \\
PD-L1 requirement & 45 & 80.8 & 75.0 & 88.3 & 90.0 & 92.6 & 92.6 & 0.0 \\
Number of arms & 46 & 78.4 & 78.2 & 86.4 & 88.8 & 91.5 & 91.5 & 0.0 \\
irAEs assessed & 44 & 28.0 & 70.8 & 0.0 & 81.4 & 89.9 & 89.9 & 0.0 \\
Combination therapy & 46 & 82.2 & 80.9 & 87.7 & 90.4 & 93.0 & 93.0 & 0.0 \\
Clinical setting & 44 & 78.8 & 77.0 & 88.3 & 84.2 & 91.3 & 91.3 & 0.0 \\
Endpoint role & 46 & 0.0 & 40.2 & 0.0 & 0.0 & 0.0 & 0.0 & 0.0 \\
PubMed link & 33 & 79.1 & 74.2 & 87.1 & 89.1 & 90.6 & 90.6 & 0.0 \\
Follow-up publication & 46 & 77.5 & 79.1 & 89.1 & 90.9 & 91.3 & 91.3 & 0.0 \\
MEK inhibitor agents & 6 & 100.0 & 100.0 & 100.0 & 100.0 & 100.0 & 100.0 & 0.0 \\
Control category & 46 & 82.2 & 78.7 & 88.3 & 88.8 & 92.5 & 92.2 & $-$0.4 \\
Combination type & 45 & 85.9 & 79.1 & 89.5 & 90.4 & 93.8 & 93.4 & $-$0.4 \\
Brand name & 46 & 62.9 & 49.4 & 87.4 & 79.4 & 89.3 & 88.8 & $-$0.5 \\
Anti-VEGF agents & 17 & 98.8 & 98.8 & 100.0 & 100.0 & 100.0 & 5.9 & $-$94.1 \\
\bottomrule
\end{tabular*}
\end{supptable}

\begin{supptable}
Components were removed one at a time. Supplementary Table~\ref{tab:s-abl-plain} gives the result for recorded-field questions, together with runs in which only part of the approved-example bank was available.
\par\vspace{6pt}
\caption{Component analysis for questions about recorded fields (300 held-out benchmark questions)}\label{tab:s-abl-plain}
\supptablefont
\begin{tabular*}{\textwidth}{@{\extracolsep{\fill}}lccc@{}}
\toprule
 & Exact match & Execution F1 & Tokens per \\
Configuration & (95\% CI) & (95\% CI) & question \\
\midrule
Full system & 62.0 (56.7--67.3) & 71.5 (66.9--75.9) & 5,859 \\
Without catalogue grounding & 64.0 (58.7--69.3) & 73.0 (68.4--77.2) & 5,703 \\
Without model decomposition & 60.0 (54.7--65.3) & 71.8 (67.2--76.0) & 3,271 \\
Without exemplar retrieval & 56.0 (50.3--61.3) & 69.9 (65.3--74.3) & 5,139 \\
Without context-field completion & 55.0 (49.7--60.3) & 69.1 (64.7--73.6) & 5,785 \\
\midrule
Example bank at 50\% & 57.7 (52.3--63.0) & 71.0 (66.3--75.2) & 5,852 \\
Example bank at 25\% & 56.3 (51.0--62.0) & 70.4 (66.0--75.0) & 5,880 \\
Example bank at 10\% & 56.0 (50.7--61.7) & 70.6 (66.1--75.1) & 5,920 \\
Example bank at 0\% & 56.0 (50.3--61.3) & 70.0 (65.3--74.5) & 5,139 \\
\bottomrule
\end{tabular*}
\end{supptable}

\begin{supptable}
Supplementary Table~\ref{tab:s-abl-hybrid} gives the corresponding analysis for derived-attribute questions. In the last row the model, rather than a query, selected the rows from the whole table.
\par\vspace{6pt}
\caption{Component analysis for questions that require a derived attribute (304-question subset)}\label{tab:s-abl-hybrid}
\supptablefont
\begin{tabular*}{\textwidth}{@{\extracolsep{\fill}}lccccc@{}}
\toprule
 & Derived-value F1 & Every row correct & Exact & Trial & Exact \\
Configuration & (95\% CI) & (95\% CI) & table & PPV & trial set \\
\midrule
Full system & 78.0 (73.7--81.8) & 66.8 (61.2--71.7) & 52.3 & 92.4 & 87.5 \\
Without program induction & 77.9 (73.7--81.8) & 66.8 (61.5--71.7) & 53.6 & 92.4 & 87.5 \\
Without output normalisation & 76.7 (72.4--80.6) & 62.2 (56.6--67.4) & 45.4 & 91.8 & 82.2 \\
Without planner & 72.0 (67.3--76.3) & 61.2 (55.9--66.4) & 47.7 & 92.4 & 87.5 \\
Model selects the rows & 67.6 (63.2--71.6) & 37.8 (32.6--42.8) & 28.9 & 77.6 & 53.0 \\
\bottomrule
\end{tabular*}
\end{supptable}

\begin{supptable}
Supplementary Table~\ref{tab:s-tasks} gives the number of clinician-style tasks that each system answered correctly, overall and for each of the six task types.
\par\vspace{6pt}
\caption{Clinician-style tasks answered correctly}\label{tab:s-tasks}
\fontsize{7.6}{9.2}\selectfont
\begin{tabular*}{\textwidth}{@{\extracolsep{\fill}}lcccccccc@{}}
\toprule
 & All (140), &  & Look-up & Count & Landscape & Enrolment & Control & PD-L1 \\
System & $n$ (\%) & 95\% CI & (40) & (40) & (24) & (23) & (9) & (4) \\
\midrule
FD-SCoPE & 140 (100.0) & 97.3--100.0 & 40 & 40 & 24 & 23 & 9 & 4 \\
DAIL-SQL style & 137 (97.9) & 93.9--99.3 & 37 & 40 & 24 & 23 & 9 & 4 \\
MAC-SQL style & 136 (97.1) & 92.9--98.9 & 39 & 38 & 24 & 23 & 8 & 4 \\
Few-shot & 136 (97.1) & 92.9--98.9 & 37 & 39 & 24 & 23 & 9 & 4 \\
Zero-shot & 134 (95.7) & 91.0--98.0 & 36 & 38 & 24 & 23 & 9 & 4 \\
C3 style & 133 (95.0) & 90.0--97.6 & 36 & 37 & 24 & 23 & 9 & 4 \\
DIN-SQL style & 132 (94.3) & 89.1--97.1 & 37 & 35 & 24 & 23 & 9 & 4 \\
Chain-of-thought & 127 (90.7) & 84.8--94.5 & 34 & 36 & 22 & 23 & 8 & 4 \\
\bottomrule
\end{tabular*}
\supptablenote{Numbers in brackets in the header are the number of tasks. The interval is the Wilson interval for the share of all 140 tasks. Task types: fact look-up for a named trial, count of trial publications, evidence landscape for a cancer type, listing by enrolment and recency, listing by a control other than chemotherapy, and listing of trials restricted to PD-L1-positive patients. Few-shot, zero-shot and chain-of-thought denote the prompting baselines.}
\end{supptable}

\begin{supptable}
Held-out benchmark questions were reworded without changing the reference answer (Supplementary Table~\ref{tab:s-robust}). Each rewording was applied only to the questions to which it was applicable.
\par\vspace{6pt}
\caption{Exact-match accuracy (\%) of FD-SCoPE on reworded questions}\label{tab:s-robust}
\supptablefont
\begin{tabular*}{\textwidth}{@{\extracolsep{\fill}}lcccc@{}}
\toprule
Rewording & Questions & Original & Reworded & Change \\
\midrule
Lower case & 300 & 62.0 & 61.7 & $-$0.3 \\
Brand name for generic drug name & 86 & 61.6 & 61.6 & 0.0 \\
Disease abbreviation & 76 & 59.2 & 60.5 & +1.3 \\
Drug-class shorthand & 81 & 67.9 & 67.9 & 0.0 \\
Typing error in a clinical term & 187 & 61.0 & 55.1 & $-$5.9 \\
\bottomrule
\end{tabular*}
\end{supptable}

\begin{supptable}
Because all systems answered the same questions, differences were also estimated on paired questions (Supplementary Table~\ref{tab:s-paired}). The last two columns count the questions on which the first system scored higher or lower than the second.
\par\vspace{6pt}
\caption{Paired differences between two systems on the same questions}\label{tab:s-paired}
\fontsize{7.5}{9.1}\selectfont
\begin{tabular*}{\textwidth}{@{\extracolsep{\fill}}lrccc@{}}
\toprule
 &  & Difference, points & Questions & Questions \\
Comparison & $n$ & (95\% CI) & better & worse \\
\midrule
\multicolumn{5}{@{}l}{\textit{Recorded-field questions: FD-SCoPE vs an alternative}} \\
\hspace{0.7em}DAIL-SQL style, exact match & 1,500 & +3.7 (+1.9 to +5.5) & 123 & 67 \\
\hspace{0.7em}DAIL-SQL style, execution F1 & 1,500 & +0.3 ($-$0.5 to +1.1) & 162 & 83 \\
\hspace{0.7em}Few-shot prompting, exact match & 1,500 & +9.6 (+7.7 to +11.5) & 193 & 49 \\
\hspace{0.7em}Few-shot prompting, exact match (Gemma-3-27B) & 1,500 & +5.1 (+2.7 to +7.6) & 227 & 150 \\
\midrule
\multicolumn{5}{@{}l}{\textit{Derived-attribute questions: FD-SCoPE vs BlendSQL style}} \\
\hspace{0.7em}Derived-value F1 & 1,500 & +4.2 (+3.2 to +5.3) & 213 & 60 \\
\hspace{0.7em}Every row correct & 1,500 & +7.9 (+6.3 to +9.5) & 140 & 22 \\
\midrule
\multicolumn{5}{@{}l}{\textit{Held-out questions: after vs before feedback}} \\
\hspace{0.7em}Derived-value F1 & 1,201 & +7.0 (+5.4 to +8.6) & 176 & 33 \\
\hspace{0.7em}Every row correct & 1,201 & +8.7 (+6.6 to +10.7) & 136 & 32 \\
\hspace{0.7em}Derived-value F1, feedback from half the trials & 1,201 & +5.4 (+3.7 to +7.1) & 172 & 63 \\
\midrule
\multicolumn{5}{@{}l}{\textit{Derived attributes: full system vs one component changed}} \\
\hspace{0.7em}Model selects the rows, F1 & 304 & +10.4 (+8.2 to +12.7) & 121 & 1 \\
\hspace{0.7em}No planner, F1 & 304 & +6.0 (+2.6 to +9.6) & 36 & 15 \\
\hspace{0.7em}No output normalisation, every row & 304 & +4.6 (+2.3 to +7.2) & 15 & 1 \\
\hspace{0.7em}No program induction, F1 & 304 & +0.1 ($-$0.6 to +1.0) & 2 & 2 \\
\midrule
\multicolumn{5}{@{}l}{\textit{Recorded fields: full system vs one component removed, exact match}} \\
\hspace{0.7em}No exemplar retrieval & 300 & +6.0 (+1.7 to +10.7) & 33 & 15 \\
\hspace{0.7em}No context-field completion & 300 & +7.0 (+2.3 to +11.7) & 35 & 14 \\
\hspace{0.7em}No model decomposition & 300 & +2.0 ($-$1.3 to +5.3) & 16 & 10 \\
\hspace{0.7em}No catalogue grounding & 300 & $-$2.0 ($-$4.0 to 0.0) & 2 & 8 \\
\bottomrule
\end{tabular*}
\supptablenote{Intervals are paired-bootstrap intervals from 10,000 resamples of questions. A positive difference favours the first system named in the group heading. F1 is derived-value F1 and every row is the share of questions with every row correct.}
\end{supptable}

\begin{supptable}
Supplementary Table~\ref{tab:s-safety} gives the outcome of the safety prompts with the scope check on and off. With the scope check off, the query guard is the only protection.
\par\vspace{6pt}
\caption{Safety set of 46 prompts}\label{tab:s-safety}
\supptablefont
\begin{tabular*}{\textwidth}{@{\extracolsep{\fill}}lcccccc@{}}
\toprule
 & & \multicolumn{2}{c}{Scope check on} & \multicolumn{2}{c}{Scope check off} & Unsafe \\
\cmidrule{3-4}\cmidrule{5-6}
Category & Prompts & Declined & Answered & Declined & Answered & statements \\
\midrule
Request to modify the table & 12 & 12 & 0 & 0 & 12 & 0 \\
Prompt injection & 10 & 10 & 0 & 0 & 9 & 0 \\
Request for clinical advice & 8 & 8 & 0 & 0 & 8 & 0 \\
Request for patient-level data & 6 & 6 & 0 & 0 & 6 & 0 \\
Valid evidence question (control) & 10 & 0 & 10 & 0 & 10 & 0 \\
\bottomrule
\end{tabular*}
\supptablenote{Answered means answered with a read-only query. Unsafe statements are statements that would modify data and were executed, summed over both runs. The table file was byte-identical before and after both runs.}
\end{supptable}

\begin{supptable}
The two language models were served with the settings in Supplementary Table~\ref{tab:s-settings}.
\par\vspace{6pt}
\caption{Language-model serving and inference settings}\label{tab:s-settings}
\supptablefont
\begin{tabular*}{\textwidth}{@{\extracolsep{\fill}}lll@{}}
\toprule
Setting & Qwen3.8-27B & Gemma-3-27B-it \\
\midrule
Numerical precision & bfloat16 & bfloat16 \\
Served context limit & 40,000 tokens & 24,000 tokens \\
Temperature & 0.0 & 0.0 \\
Top-$p$ & 1.0 & 1.0 \\
Thinking mode & Disabled & Not applicable \\
JSON-constrained output & Enabled & Enabled \\
Token log-probabilities & Enabled & Disabled \\
Concurrent requests per process & 64 & 24 \\
Serving software & vLLM 0.27.1 & vLLM 0.27.1 \\
Hardware & NVIDIA H200 & NVIDIA H200 \\
\bottomrule
\end{tabular*}
\end{supptable}

\begin{supptable}
The alternative approaches were configured as listed in Supplementary Table~\ref{tab:s-baselines}.
\par\vspace{6pt}
\caption{Settings of the alternative approaches}\label{tab:s-baselines}
\supptablefont
\begin{tabular*}{\textwidth}{@{\extracolsep{\fill}}p{0.27\textwidth}p{0.68\textwidth}@{}}
\toprule
Approach & Settings \\
\midrule
Zero-shot prompting & One call; schema and up to 24 observed values per field; 400 new tokens. \\
Few-shot prompting & As zero-shot, with four fixed expert examples; 400 new tokens. \\
Chain-of-thought prompting & Reasoning scaffold with worked examples; 2,500 new tokens. \\
DIN-SQL style & Four calls: schema linking (400 tokens), difficulty classification (10), generation (500) and self-correction (1,200); the corrected query is kept only if it executes. \\
DAIL-SQL style & Eight examples chosen by similarity of the question with literals masked, most similar last; schema given as a table definition; 500 tokens. \\
C3 style & Zero-shot prompt with clear layout and tips; ten samples at temperature 0.7; the result shared by most executable samples is returned. \\
MAC-SQL style & Selector (300 tokens), decomposer (1,200) and refiner (600) roles; up to three refinement rounds on error or empty result. \\
BlendSQL style & Plan with a filter, a field to map and a question per value (500 tokens); the model answers over distinct values in batches of 40 (2,000 tokens). \\
Code agent (EHRAgent style) & The model writes pandas code (1,500 tokens) run in a restricted subprocess with a 60-second limit and a read-only database; up to three repair rounds. \\
Row-retrieval question answering & Rows embedded with all-MiniLM-L6-v2; the 40 most similar rows are given to the model; 3,000 tokens. \\
Whole-table prompting & The whole table in one prompt (about 28,000 prompt tokens per question); 3,000 new tokens. \\
\bottomrule
\end{tabular*}
\supptablenote{All approaches used the same model, table and read-only database, at temperature 0 unless stated.}
\end{supptable}


\subsection*{Supplementary Methods: complete model prompts}

The following templates reproduce the system and user prompts used by
FD-SCoPE and the prompting comparators. Expressions enclosed in double
braces denote values inserted programmatically at runtime. These values
included the clinical question, database schema, observed values, grounded
predicates, retrieved exemplars, selected rows and intermediate outputs.

\subsubsection*{FD-SCoPE prompts}

\paragraph{Question routing.}

\begin{promptbox}{FD-SCoPE Prompt 1: Question routing}
SYSTEM

[stage:route] You are the routing stage of a clinical-trials
question-answering system over one SQLite table.
Decide whether a question can be answered with a plain SQL query
(filtering or projecting stored columns) or whether it asks for a
LATENT value that is not stored and must be derived per row
(classify, bucket, normalise, extract, map, flag, convert,
construct).

USER

<<schema>>
{{schema}}
<</schema>>

<<question>>
{{question}}
<</question>>

Return ONLY JSON:
{
  "kind": "sql" | "hybrid",
  "filter_text": "the part of the question that selects rows
                  (cohort filters), rewritten as a standalone
                  question",
  "derivation_instruction": "the per-row derivation asked for
                             (empty when kind = sql)",
  "answer_type": "string | boolean | number | list",
  "source_column": "the stored column from which the derived value
                    is computed (empty when kind = sql)"
}
\end{promptbox}

\paragraph{Question decomposition.}

\begin{promptbox}{FD-SCoPE Prompt 2: Question decomposition}
SYSTEM

[stage:decompose] You are an expert SQL query decomposition
assistant for the clinical_trials table. You decompose a question
into atomic, retrieval-friendly sub-questions that each map to
exactly one SQL WHERE predicate (column, operator, value).
You never write SQL and never output projection instructions.

USER

<<schema>>
{{schema}}
<</schema>>

{{routing_rules}}

<<question>>
{{question}}
<</question>>

Rules:
- One sub-question per constraint; preserve every constraint
  exactly (cancer type, drug, class, phase, control arm, year,
  arms, follow-up ...).
- Use explicit operators when implied: =, >=, <=, >, <, !=.
- Use the canonical column name and canonical literal from the
  schema above.
- Phrase each sub-question as:
  Which trials have <column> <operator> <value>?
- Return between 1 and {{max_subquestions}} sub-questions.
- If the question has no filter, return an empty list.

Good example:

Question: Which colorectal trials had 3 or more arms and used a
multikinase inhibitor as a control?

{
  "decomposed_queries": [
    "Which trials have Cancer type = 'Colorectal'?",
    "Which trials have Number of arms >= 3?",
    "Which trials have Control arm = 'Multikinase inhibitor'?"
  ]
}

Bad example:

{
  "decomposed_queries": [
    "Find relevant colorectal studies.",
    "Get trials with enough arms.",
    "Show NCT and Author."
  ]
}

Output ONLY valid JSON:

{"decomposed_queries": ["...", "..."]}
\end{promptbox}

\paragraph{SQL synthesis.}

\begin{promptbox}{FD-SCoPE Prompt 3: SQL synthesis}
SYSTEM

[stage:synthesis] You are an expert SQLite NL2SQL generator for
the clinical_trials table. You write ONE read-only SELECT
statement that answers the question exactly, using only the
schema, grounded predicates and retrieved expert-approved
exemplars as structural hints.

USER

<<schema>>
{{schema}}
<</schema>>

{{allowed_values}}

{{routing_rules}}

Evidence priority (highest to lowest):
1. The parent question
2. The grounded predicates
3. The decomposed sub-questions
4. Exemplar SQL as structural hints only

Predicate budget:
The WHERE clause contains exactly the constraints the question
asks for. Never copy literals from unrelated exemplars and never
add extra filters.

Projection:
If the question names the fields it wants, SELECT exactly those
fields and nothing else. A field that is only listed as wanted
output is never a filter. Otherwise, SELECT the identifier columns
"NCT", "Author", "Year", and
"Originial publication or Follow-up", followed by the columns used
in the filters and closely related context columns.

Always double-quote column names exactly as in the schema and
single-quote string literals. End with a semicolon.

<<question>>
{{question}}
<</question>>

<<decomposition>>
{{decomposition}}
<</decomposition>>

<<grounded_predicates>>
{{grounded_predicates}}
<</grounded_predicates>>

<<exemplars>>
{{exemplars}}
<</exemplars>>

{{extra_instructions}}

Return ONLY JSON:

{"sql": "SELECT ... ;"}
\end{promptbox}

\paragraph{SQL repair.}

\begin{promptbox}{FD-SCoPE Prompt 4: SQL repair}
SYSTEM

[stage:sql_repair] You repair a SQLite SELECT statement for the
clinical_trials table so that it executes and still answers the
question. Change as little as possible: fix column names, literals
or syntax using the schema; never add or remove constraints.

USER

<<schema>>
{{schema}}
<</schema>>

<<question>>
{{question}}
<</question>>

<<sql>>
{{sql}}
<</sql>>

<<error>>
{{error}}
<</error>>

Return ONLY JSON:

{"sql": "SELECT ... ;"}
\end{promptbox}

\paragraph{Fallback row selection.}

\begin{promptbox}{FD-SCoPE Prompt 5: Fallback row selection}
SYSTEM

[stage:select_rows] You are the row-selection stage of a
clinical-trials table reasoning system. Identify which visible
rows satisfy the question's filters. Do not derive the final
answer.

USER

Use only the visible table. The column "__rowid__" uniquely
identifies each row.

Return ONLY JSON:

{
  "selected_row_ids": [1, 2, 3],
  "selection_reason": "...",
  "needs_derivation": true
}

<<question>>
{{question}}
<</question>>

<<headers>>
{{headers_json}}
<</headers>>

<<table_csv>>
{{table_csv}}
<</table_csv>>
\end{promptbox}

\paragraph{Hybrid derivation planning.}

\begin{promptbox}{FD-SCoPE Prompt 6: Hybrid derivation planning}
SYSTEM

[stage:plan] You are the planner of a hybrid querying system over
a clinical-trials table. The cohort rows have already been
selected with SQL. Your job is to explain how to derive, for EACH
row, the latent value the question asks for from the visible
source column and, when the rule is deterministic, to express it
as a small Python expression.

USER

<<question>>
{{question}}
<</question>>

<<instruction>>
{{instruction}}
<</instruction>>

<<source_column_hint>>
{{source_column_hint}}
<</source_column_hint>>

<<answer_type_hint>>
{{answer_type_hint}}
<</answer_type_hint>>

<<headers>>
{{headers_json}}
<</headers>>

<<candidate_preview>>
{{candidate_preview_json}}
<</candidate_preview>>

Guidance:
- inferred_source_column must be one of the visible headers.
- Prefer the source-column hint unless the preview shows it is
  wrong.
- derivation_rules must contain precise and exhaustive rules
  mapping source values to outputs.
- normalization_rules must define the exact output labels and
  format.
- Represent booleans as true or false.
- Represent lists as JSON arrays.
- program is OPTIONAL.
- A program must be one Python expression over `v`, representing
  the source value, and `row`, representing a dictionary of the
  visible columns.
- Permitted operations include string methods, re, json, len, int,
  float, round, any, all, sorted, min, max, comparisons,
  conditional expressions and list comprehensions.
- Return null when the value cannot be derived.
- Supply a program only when the rule is fully deterministic.
- Never use imports or unlisted names.

Return ONLY JSON:

{
  "inferred_source_column": "...",
  "derived_field_description": "...",
  "answer_type": "string|boolean|number|list",
  "relevant_columns": [
    "__rowid__", "NCT", "PubMed ID", "Trial name", "..."
  ],
  "derivation_rules": ["..."],
  "normalization_rules": ["..."],
  "program": "<python expression or null>",
  "notes": "..."
}
\end{promptbox}

\paragraph{Row-level derivation.}

\begin{promptbox}{FD-SCoPE Prompt 7: Row-level derivation}
SYSTEM

[stage:derive] You are the executor stage of a hybrid querying
system over a clinical-trials table. For every candidate row you
output the derived value requested by the question, following the
planner's rules exactly.

USER

<<question>>
{{question}}
<</question>>

<<instruction>>
{{instruction}}
<</instruction>>

<<plan>>
{{plan_json}}
<</plan>>

<<rows>>
{{rows_json}}
<</rows>>

Rules:
- Return exactly one prediction per row in <<rows>>, using its
  integer "__rowid__" as table_row_id.
- Derive the answer from the row's source value, represented by
  "source_value", and the visible context.
- Follow the normalization rules.
- Use null when the value cannot be derived.
- Booleans must be JSON true or false.
- Lists must be JSON arrays.
- Do not add explanations.

Return ONLY JSON:

{
  "predictions": [
    {"table_row_id": 1, "answer": ...}
  ]
}
\end{promptbox}

\paragraph{SQL back-translation.}

\begin{promptbox}{FD-SCoPE Prompt 8: SQL back-translation}
SYSTEM

[stage:back_translate] You rewrite a SQLite query over the
clinical_trials table into the natural question a clinician would
ask, keeping the intent identical, and then decompose it into one
sub-question per WHERE predicate.

USER

<<schema>>
{{schema}}
<</schema>>

<<original_question>>
{{original_question}}
<</original_question>>

<<sql>>
{{sql}}
<</sql>>

Rules:
- Preserve every filter and threshold.
- Do not mention SQL, tables or column names in the question.
- Keep numbers as digits.
- Prefer explicit disease and drug names.

Return ONLY JSON:

{
  "question": "...",
  "decomposed_queries": [
    "Which trials have <column> <operator> <value>?",
    "..."
  ]
}
\end{promptbox}

\paragraph{Derivation-program induction.}

\begin{promptbox}{FD-SCoPE Prompt 9: Derivation-program induction}
SYSTEM

[stage:induce_program] You write a single Python expression that
reproduces a clinician's derivation rule over one column of a
clinical-trials table. The expression is executed for every row
with the variables `v`, representing the source cell as a string,
and `row`, representing a dictionary of the row.

USER

<<instruction>>
{{instruction}}
<</instruction>>

Source column: "{{source_column}}"
Expected answer type: {{answer_type}}

Clinician-approved examples
(source value -> expected output, JSON):

<<examples>>
{{examples_json}}
<</examples>>

{{feedback}}

Requirements:
- Return ONE Python expression.
- Do not use statements, lambdas or imports.
- Permitted operations include string methods, re, json, len, int,
  float, str, round, any, all, sorted, min, max, list, dict, set,
  tuple, bool, abs, sum, enumerate, zip, range, comparisons,
  conditional expressions, comprehensions and inline
  dictionaries.
- Reproduce expected outputs exactly, including spelling, case,
  list order, booleans and null values.
- Prefer the simplest rule that explains all examples.
- For lookup tasks, construct a dictionary from the examples and
  apply it to components of `v`.
- Do not enumerate long lists of values absent from the examples.
- Keep the expression under 1,500 characters.
- Generalize sensibly to unseen values when an evident rule
  exists.

Answer in exactly this format:

<program>
the Python expression
</program>

<summary>
one sentence a clinician can verify
</summary>
\end{promptbox}

\subsubsection*{Comparator prompts}

The following prompts were used for the zero-shot, few-shot and
chain-of-thought natural-language-to-SQL comparators. The schema,
observed values and evaluated question were inserted at runtime. Prompts of the
method baselines are provided with the code.

\paragraph{Zero-shot prompting.}

\begin{promptbox}{Comparator Prompt 1: Zero-shot prompting}
You are an expert natural language to SQL generator. Write one
SQLite SELECT query over "clinical_trials".

Rules:
- Output only the SQL query (no prose, no markdown).
- The SQL MUST start with SELECT.
- Use only columns from this table schema: {SCHEMA}
- Put double quotes around EVERY column name exactly as in the
  schema.
- Use single quotes for string literals.
- Include all constraints from the question.

{OBSERVED_VALUES: up to 24 values per database field}

Question: {QUESTION}
Generate SQL:
\end{promptbox}

\paragraph{Few-shot prompting.}

The few-shot baseline used four fixed clinician-approved demonstrations
representing disease and treatment constraints, combination therapy, comparator
selection and biomarker eligibility. The same four demonstrations were used for
every question and model. One representative demonstration is shown below.

\begin{promptbox}{Few-shot prompt: representative demonstration}
The zero-shot instruction, schema and observed values were supplied first.

Examples:

Question: Find renal cell trials of nivolumab given in
combination therapy.

SQL:
SELECT "NCT", "Author", "Year",
       "Originial publication or Follow-up",
       "Name of ICI", "Monotherapy/combination",
       "Cancer type", "Type of combination", "Trial phase"
FROM clinical_trials
WHERE "Cancer type" = 'Renal cell'
  AND "Name of ICI" = 'Nivolumab'
  AND "Monotherapy/combination" = 'Combination';

Additional demonstrations addressed:
- PD-L1 bladder trials beginning in or after 2018;
- melanoma trials using vaccine as the control arm; and
- head and neck trials requiring a non-PD-L1 biomarker.

Question: {QUESTION}
Generate SQL:
\end{promptbox}

\paragraph{Chain-of-thought prompting.}

The chain-of-thought baseline used the following reasoning scaffold. The
runtime prompt also contained one positive and one negative worked example;
the complete examples are provided in the machine-readable prompt file.

\begin{promptbox}{Chain-of-thought prompt: reasoning scaffold}
You are an expert natural language to SQL generator.
Query cancer drug trial studies from the "clinical_trials" table.

Use only columns from this schema:
{SCHEMA}

Return two blocks in this order:

<THINKING>
1. Normalize clinical terms to stored schema values.
2. Identify the retrieval intent.
3. Map every constraint to an exact database column.
4. Select the comparison operators and AND/OR relationships.
5. Select the identifiers, contextual fields and requested
   output fields.
</THINKING>

<SQL>
A single read-only SQLite SELECT statement.
</SQL>

SQL rules:
- The query must begin with SELECT.
- Use only columns present in the supplied schema.
- Double-quote column names.
- Single-quote string literals.
- Preserve every constraint in the question.
- Do not include text outside <THINKING> and <SQL>.

{OBSERVED_VALUES: up to 24 values per database field}

Question: {QUESTION}
\end{promptbox}


\bibliography{references}

\end{document}